\documentclass{ieeeaccess}
\usepackage{cite}
\usepackage{amsmath,amssymb,amsfonts}
\usepackage{algorithmic}
\usepackage[ruled,vlined]{algorithm2e}
\usepackage{graphicx}
\usepackage{textcomp}
\def\BibTeX{{\rm B\kern-.05em{\sc i\kern-.025em b}\kern-.08em
    T\kern-.1667em\lower.7ex\hbox{E}\kern-.125emX}}

\ifCLASSOPTIONcompsoc
\usepackage[caption=false, font=normalsize, labelfont=sf, textfont=sf]{subfig}
\else
\usepackage[caption=false, font=footnotesize]{subfig}

\begin{document}
\history{Date of publication xxxx 00, 0000, date of current version xxxx 00, 0000.}
\doi{10.1109/ACCESS.2017.DOI}

\title{An Extension of BIM Using AI: a Multi Working-Machines Pathfinding Solution}
\author{\uppercase{Yusheng Xiang}\authorrefmark{1,2,3}, \IEEEmembership{Member, IEEE},
\uppercase{Kailun Liu}\authorrefmark{2}, \IEEEmembership{Student Member, IEEE},
\uppercase{Tianqing Su}\authorrefmark{1,4}, \IEEEmembership{Member, IEEE},
\uppercase{Jun Li}\authorrefmark{1,5},
\IEEEmembership{Member, IEEE},
\uppercase{Shirui Ouyang}\authorrefmark{2},
\uppercase{Samuel S.  Mao}\authorrefmark{1,3},
\uppercase{Marcus Geimer}\authorrefmark{2}
}
\address[1]{Elephant Tech LLC, Shenzhen, China}
\address[2]{Institute of Mobile Machines, Karlsruhe Institute of Technology, Karlsruhe, 76131 Germany (e-mail: marcus.geimer@kit.edu)}
\address[3]{Department of Mechanical Engineering, University of California at Berkeley, Berkeley, CA, 94720, USA (e-mail: ssmao@berkeley.edu) }
\address[4]{Guanghua School of Management, Peking University, Beijing, 100081, China 
(e-mail:  t.su1991@gmail.com)}
\address[5]{Center of AI, Faculty of Engineering
and Information Technology, University of Technology Sydney, Ultimo,
NSW, Australia
(e-mail: jun.li@uts.edu.au) }

\markboth
{Author \headeretal: Preparation of Papers for IEEE TRANSACTIONS and JOURNALS}
{Author \headeretal: Preparation of Papers for IEEE TRANSACTIONS and JOURNALS}

\corresp{Corresponding author: Yusheng Xiang (e-mail: yusheng.xiang@partner.kit.edu).}

\begin{abstract}
Multi working-machines pathfinding solution enables more mobile machines simultaneously to work inside of a working site so that the productivity can be expected to increase evolutionary. To date, the potential cooperation conflicts among construction machinery limit the amount of construction machinery investment in a concrete working site. To solve the cooperation problem, civil engineers optimize the working site from a logistic perspective while computer scientists improve pathfinding algorithms' performance on the  given benchmark maps. In the practical implementation of a construction site, it is sensible to solve the problem with a hybrid solution; therefore, in our study, we proposed an algorithm based on a cutting-edge multi-pathfinding algorithm to enable the massive number of machines cooperation and offer the advice to modify the unreasonable part of the working site in the meantime. Using the logistic information from BIM, such as unloading and loading point, we added a pathfinding solution for multi machines to improve the whole construction fleet's productivity. In the previous study, the experiments were limited to no more than ten participants, and the computational time to gather the solution was not given; thus, we publish our pseudo-code, our tested map, and benchmark our results. Our algorithm's most extensive feature is that it can quickly replan the path to overcome the emergency on a construction site.

\end{abstract}

\begin{keywords}
Smart working site, Multi agents pathfinding, Conflict based searching, Building Information Model, Construction site productivity, Huoshenshan hospital project, Path planning
\end{keywords}

\titlepgskip=-15pt

\maketitle

\section{Introduction}
\label{sec:introduction}
\PARstart{A}{lthough} many achievements in construction machines with respect to productivity and safety, humans' pursuit of even higher productivity and better safety never stops. In the past decades, civil engineers and construction-industry-related software engineers introduce the Building Information Model (BIM) \cite{Borrmann.2018} as a powerful tool to increase productivity and safety performance whilst reduce the project cost by means of digital technology. In general, BIM provides the 3D or more than 3D model of the construction projects and even the installation sequence of the components to avoid mistakes during the construction stage.  With the maturity of BIM, this software and process are adopted for many large and especially complicated construction projects worldwide \cite{Barlish.2012} since the mistakes during the real construction process cause much more severe consequences than those in virtual engineering. Also, BIM is considered as a lifelong software, contributing to not only the early construction phase \cite{Borrmann.2018, Liu.2017} but also the time after the construction projects are finished \cite{Volk.2014}. Despite to the preliminary cost for training corresponding laborers and model building in a computer, BIM is a necessary tool for at least large construction projects becomes a consensus. However, although current BIM software defines the start and endpoints where the material should be transported, concrete paths guiding the trucks to accomplish the goal are not given. Or more generally, an algorithm that determines the paths of the participants in the working site so that they can move to their destination without collision and hesitation is still developing. 

\Figure[ht!](topskip=0pt, botskip=0pt, midskip=0pt)[width=7in]{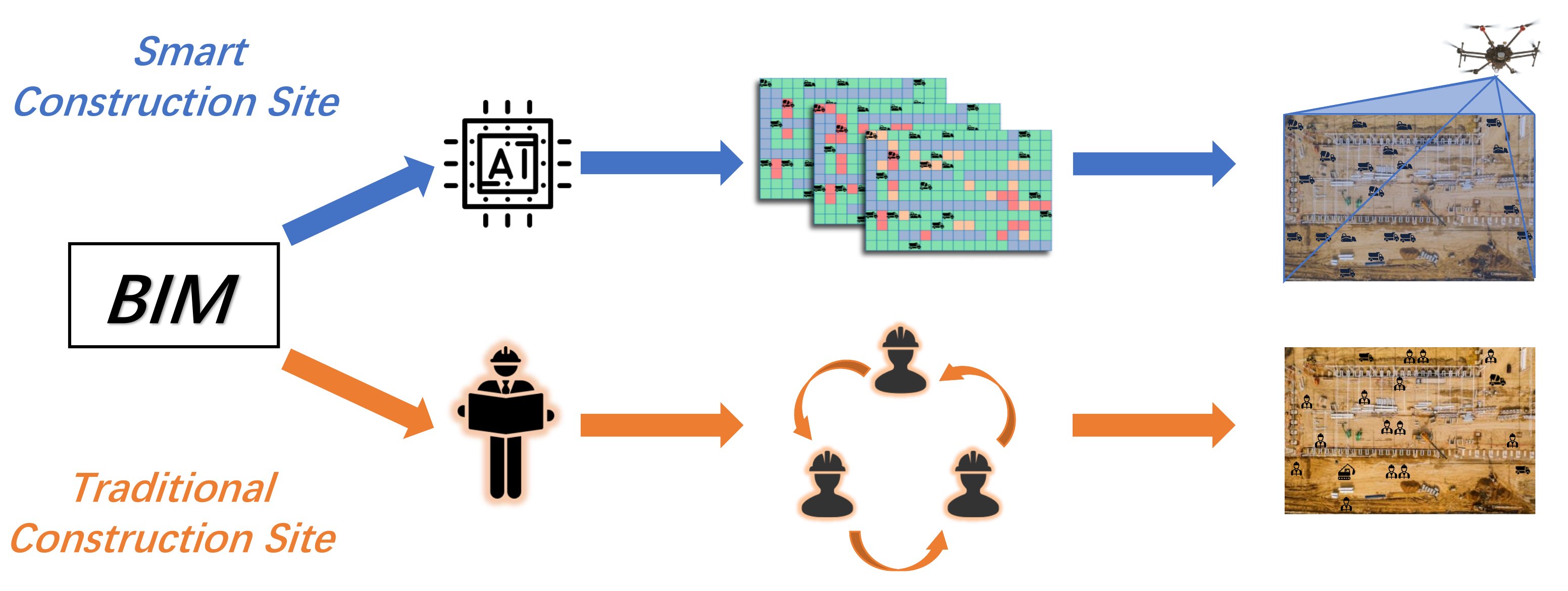}
{Overview of the smart construction site concept. We introduce to use  artificial intelligence to unify the scheduling of construction machinery on construction sites. Maximize the use of construction machinery on specific construction sites by avoiding conflicts among construction machinery and thus ultimately improve the productivity of the construction sites. \label{fig:motivation}}

As shown in Fig. \ref{fig:motivation}, one motivation to combine these path planning algorithms with BIM can be described as determining the construction machines' travel path so that the machines can be expected to move faster and denser without hesitation. To accomplish higher productivity and better safety in the path planning fashion, some researchers calculate and display efficient paths by a given construction site \cite{Song.2019, soltani.2002} whereas others optimize the construction site layout considering the high productivity of the path \cite{Khalafallah.2011, Yahya.2014}. However, most of the current solutions about path planning on a construction site are mainly focusing on individual units, i.e., the interaction and path conflict of the different machines or other agents are ignored. Consequently, the working sites' spatial and time utilization is limited. Inspired by warehouse logistics solution \cite{Ma.2019}, where a lot of robotics are working with the commands from path planning algorithms in the meantime and thus achieve considerably higher transport efficiency as a fleet. We envision the multi-agents pathfinding (MAPF) solution can also provide evolutionary to the construction industry. A persuasive instance to show the benefit of introducing MAPF solution into a working site is the construction site for the famous hospital, namely Huoshenshan, in Wuhan during the coronavirus 2019 outbreak. Invested an extraordinary amount of working machines and human cooperators, the construction project was finished at an unprecedented speed, through only manually coordinate to avoid conflicts among machines. Apparently, the economic cost of running such a construction site can be quite expensive due to experienced workers. Also, since the MAPF problem is NP-hard, computer algorithms can surely have a better performance concerning a series of optimization objects, such as shorter moving distance and realtime performance. In light of that, we propose a MAPF solution for working machines as an extension of the BIM system so that more machines can work simultaneously and thereby achieve better holistic productivity.  

The aim of this paper is to extend the current BIM software with Artificial Intelligence (AI) path-planning algorithms in order that more machines can work simultaneously since the paths are calculated to avoid collisions.  Because the start points and endpoints can be given directly in the BIM system, we devote ourselves to the implementation level, i.e., how exactly the machines should achieve their goal set given by the BIM system.  We envision that the case in Wuhan will be a normal case in the future by utilizing AI and IoT \cite{Xiang.2020f, Xiang.2020} technology.

The main contributions of this paper can be sum up as the following points:
\begin{itemize}
    \item We introduced a novel MAPF solution to guide a fleet of mobile machines to work simultaneously inside a working site and to give suggestions to optimize the construction site layout. 

    \item The approach we proposed can always find a feasible solution on weighted maps considering the agents' priority in a predefined short period to deal with emergencies.
    
    \item We extended the cutting-edge MAPF solution with a bidirectional searching method for the initial search of the best path, which shall be principally faster.
    
    \item Our method can be added to BIM software to make up for its lack of path planning.
    
    \item We benchmark the performance of our MAPF solution for the mobile machines with respect to the solution found time and cost to reach their goal on the given maps.

\end{itemize}

The rest of this paper is organized as follows. Section II briefly introduces the prerequisite and background knowledge in fields of BIM, path planning methods for construction machines and robotics to understand this paper quickly. Next, the existing problems are illustrated in Section III. After that, in Section IV, we describe the setup of our MAPF approach, including the map to give the start and goal position, low level for individual pathfinding, and high level for conflicts solving. Then, we show the experiments setup. Followed by section VI, we show our approach's performance by testing on some maps based on real construction sites. Finally, Section VII summarizes the advantages of our approach, and Section VIII gives conclusions and envisions the outlook.

\section{related works}

\subsection{Building information modelling}

Building Information Modelling (BIM) is a 3D model-based information management process in the field of Architecture, Engineering, and Construction (AEC) that facilitates efficient design and construction processes and inter-organizational collaboration \cite{Whitlock.2018}. There is a lot of BIM-based software: Autodesk Revit Building (Revit), ArchiCAD, Bentley, and SolidWorks \cite{Woo.2006}. By building the whole project virtually before physical construction begins, construction sequencing is determined, including material ordering, fabrication, and delivery schedules for all building components, etc. Therefore, conflict, interference, and collision are avoided in the early stage, contributing to improved site efficiency and reduced cost \cite{Azhar.2011}. As the function inside BIM increases, such as scheduling, virtual reality \cite{Rahimian.2020}, and logistic management \cite{Whitlock.2018}, it has been extended to 4 or more than 4 dimensional model. Nowadays, the research about BIM is prosperous. Combining AI and IoT into the BIM system is considered the next potential boom for BIM systems. Survey papers about that can be found in \cite{ Sacks.2020, Tang.2019}.

In order to automate the whole construction site and compute the optimal path for the heavy machines, logistic information is vital, which demonstrates which materials should be placed in which location at which time in the right quantity. Logistics management in construction involves the strategic storage, handling, transportation and distribution of resources, as well as planning of a building site’s layout \cite{Sullivan.2011}. Whitlock has proposed a desktop approach to adopt BIM for construction logistics management \cite{Whitlock.2018}.  Such logistic information, for instance, unloading points, on-site arrangements-logistics layouts, which are generated at the outset of the pre-construction process, can be used as the input data for the path planning.  

\subsection{path planning for construction machines}

In a construction site, there are usually multiple machines working simultaneously in a definite area with given assignments. Therefore, coordinated construction logistics can definitely increase productivity, decrease material usage, and guarantee workers' health and safety \cite{HedborgBengtsson.2019}. To date, there is plenty of researchers proposed their solution for the logistical problem inside construction sites at diverse levels, i.e., path planning and motion control. In this section, we summarize the previous research about moving paths inside of construction sites. 

In the construction industry, the earth-moving sector is among the pioneers in adopting new sensing and information technologies \cite{Azar.2017}, such as bulldozer \cite{Hirayama.2019, Hirayama.2019b}, and grading machines \cite{Sun.2018}.  Given two points A and B on a construction site, the objective is to determine the shortest path from A to B maintaining a safe distance from obstacles. The approach proposed by Kim is a path-planning method for a mobile construction robot to find a continuous collision-free path from the initial position of the construction robot to its target position by improved Bug-based algorithm \cite{Kim.2003}. The algorithm can work with the disturbance of static and dynamic objects. Obviously, the performance of the approach is based on the accuracy of the sensors. At that time, the methods to acquire site information were still immature. Hence, the spatial model supporting path planning in a partially known and partially unknown environment was brought forward by Lee. Accordingly, the spatial model provides the domain for finding an efficient path in a construction site through the use of an algorithm that combines a shortest path algorithm and a dynamic path-planning algorithm. This approach differs from existing path-planning approaches that assume the construction site is totally known or totally unknown. Thus, problems associated with managing a changing construction environment and ignorance of designated roadway networks are overcome \cite{Lee.2004}. In the same year, Soltani has compared the performance of different methods for the path planning inside construction sites \cite{Soltani.2004}, such as Dijkstra \cite{Dijkstra.1959},  $A^*$ \cite{hart.1968}, and Genetic Algorithm (GA) \cite{Holland.1992}, by evaluating comprehensive multi objects, e.g., site layout representation, distance formulation, hazard zone modeling, and visibility calculations. Also, the author points out the use of CCTV cameras should be considered to enhance site security \cite{Soltani.2004}. Although the simulation results show that the GA has the best performance and the other two algorithms have quite similar results, we conjecture that it might be ascribed to their maps, which are quite easy and do not include difficult obstacles such as bottlenecks. Fairly recently, Song tried to integrate some path planning algorithms into BIM system \cite{Song.2019}. The basic idea is to determine the path to transport the materials at the very beginning phase, i.e., the construction site design phase.  Also, the study verified the demand for the introduction of path planning by survey questions. The shortcoming of this approach is that the interaction of other participants inside construction sites during the construction project was not taken into account. So far, the aforementioned studies focus on a path for one machine inside the construction site. In contrast, the following research shows solutions for multi machines working simultaneously on a construction site. An influential study about the path planning on the construction site is from Cheng published in 2012 \cite{Cheng.2012}. The objective of the paper is to provide the n best and safest paths between two points in a work area while maintaining a safe distance from identified obstacles. Here the approach proposed in the paper uses Dijkstra algorithm and solves the path of different participants in sequence. Due to the limited recognition distance of ultra-wideband sensors, the usage of this approach might not suitable for huge working sites. Also, since the paths of agents are calculated one by one, the computation efficiency is not ideal from today's point of view in 2020. 4 years  after Cheng's study, the research from Bohacs shows the difficulties of path planning for a construction site. Also, they use $A^* $ as basic and develop an algorithm to let limited machines can cooperate without collision \cite{Bohacs.2016} within a small map. Concretely, they showed the demo about 3 machines in a $10 \times 10$ grid map. As the flow chart in their paper shows, the algorithm depends on the condition statements. As a result of that, it cannot perform with all the dispensable computation effort of the computer.

As the development of information technology, Štefanič provides an overview of emerging smart construction applications in areas such as construction monitoring, construction site management, safety at work, early disaster warning, and resources and assets management \cite{Stefanic.2019}. Also, Tumer describes the future construction site utilizing industry 4.0 \cite{Turner.2021}. Without a doubt, a future working site should be fully benefited from AI \cite{Xiang.2020h, Xiang.2020c, Xiang.2020g}, automation technology \cite{Kurinov.2020}, Simultaneous Localization and Mapping (SLAM) \cite{Xiang.2020b}, and Internet of Things (IoT) \cite{Zhou.2019, Xiang.2020f, Zhou.2017, Xiang.2020}. Therefore, the uncertainty degree of construction sites is reducing, and thus we consider the construction sites as known in our research.

Based on the strict literature review, $A^* $ is the most developed and latest algorithm to solve construction sites' path tasks. It combines both step-cost calculation from the Dijkstra algorithm and feedback step from the genetic algorithm. However, as the number of machines increases, naive $A^* $ might not be suitable to solve the MAPF problem due to algorithm complexity. Thus, it is necessary to explore the SOTA solutions in the field of mobile robotics in order to find a more appropriate solution.

\subsection{MAPF for mobile robotics}

For the single agent, the planning task can be described as finding the lowest cost from the starting point to the targeting point. By using Heuristic Search, e.g., $A^*$ Algorithm, such problem can be better solved. However, naive  $A^*$ Algorithm only considers that all the obstacles are static, which is the ideal assumption in the pathfinding problem \cite{Stern.2019}. In contrast, to solve the MAPF problem, we need to consider the other participants in the map, i.e., the dynamic obstacles also affect the optimal path of an individual agent. 

MAPF for mobile robotics is both a well-studied and dynamic developing topic. The usage scenarios of MAPF and their corresponding algorithm are diverse, such as warehouse \cite{Ma.2019}, computer games \cite{Sturtevant.2012}, and autonomous driving in intersection \cite{Svancara.2019}. Until the end of 2020, although the concept of reinforcement learning is attracting more and more attention, current influential research about the shortest path and MAPF is mainly based on graph theory. To date, there is no universal solution for all kinds of pathfinding problems; thus, algorithms with different time and space complexities are proposed \cite{Madkour.2017}. 

Based on the survey paper from Felner \cite{Felner.2018b}, we know that no algorithm dominates all others in all circumstances. There is a tradeoff between high-quality path solutions and realtime performance. The mainstream MAPF solution can be classified into search-based solvers and rule-based solvers. The former intends to find the best solution or near-optimal solutions, whereas the latter can run much faster, however, produce far away from optimal solutions.  Of course, some compromised solutions combined two ideas together, namely, hybrid approaches. Another type of solution, namely reduction-based optimal solvers, focusing on reducing MAPF problems into some problems, such as the Constraint Satisfaction Problem (CSP), with a well-known solution. Since this approach usually only aims at the makespan tasks, we will not go much deeper in this approach. To date, the most influential solutions for MAPF are Conflict Based Search (CBS) and its variants due to their widely used real-world applications. 

Sharon proposes Conflict Based Search (CBS), combining both advantages from coupled and decoupled approaches. Although the pathfinding process is strictly single-agent searches, it can guarantee to offer optimal results, unless the one variants which deliberately provide a suboptimal solution for the purpose of realtime performance. As the authors introduce, CBS adopts a two-level structure, where the high-level search can be described as a Constraint Tree (CT), including every constraint. Then, the lower level finds the concrete path for each agent individually with the information from the higher level. The brilliance of this design is that the search process is not more exponential in the number of participants but exponential in the amount of conflicts encountered during the pathfinding process. 

Since CBS tries to find the optimal solution and thereby cause a relatively longer runtime. In improved CBS algorithm \cite{Boyarski.2015}, Boyarski summaries two methods to reduce the runtime. Concretely, it firstly adopts the Meta-Agent CBS (MA-CBS) \cite{Sharon.2012}, which merges multi-agents together and handles them as a large agent. Moreover, it uses bypass improvement, which encourages one of the agents to find an alternative path instead of performing a split at the high level. As mentioned by Boyarski, the bypass concept successfully avoid the unnecessary generate the new nodes in the CT \cite{Boyarski.2015}. Since the bypass concept only tries to find the solution from the path with the same cost as the one that shall be replaced, the optimality cannot be harmed. In the high-level search, Felner suggests adding heuristics into CBS so that the conflicts are not arbitrarily chosen \cite{Felner.2018}. After that, Li found the improved heuristics to guide the high-level search \cite{Li.2019}. Also, she introduced the CBS with disjoint splitting  \cite{Li.2019b}. The main contribution of CBS with disjoint splitting is the novel terminology of positive constraints forcing the $a_i$ to be at $v$ at timestep $t$. In this fashion, CBS with disjoint slitting reduces the amount of unnecessarily expanding the CT. In addition, some improvements, such as Lazy CBS, which avoids the behavior that CBS resolves the same conflicts between the same pairs of agents many times owing to lack of connection among subproblems \cite{Gange.2019}. Hönig proposed an approach called  Conflict-Based Search with Optimal Task Assignment (CBS-TA) \cite{Honig.2018}. The improvement is mainly because it creates the forest on demand. Solving MAPF optimally is proven to be NP-hard, so CBS and all other optimal solvers do not scale up. Alternatively, Barer proposed a suboptimal variant of the CBS algorithm \cite{Barer.2014} so that the problem can be solved suboptimally but much faster. 

To sum up, naive CBS is an optimal pathfinding solution that is based on graph theory. The time consumption of the algorithm mainly depends on the conflicts occurring among the agents since they increase the nodes in the high-level tree. Its performance on bottlenecks and corridors is better, whereas the performance on open space can be worse than enhanced $A^*$. Thus, CBS's variants are focusing on reducing the nodes in the CT and therefore let CBS has a higher success rate in general. Note that in this paper, we use the same terminology as Stern's research to avoid misunderstanding; however, some mathematical descriptions might be adjusted.

\section{Problem statement}

Although the path planning problem has been attracted engineer's attention, the proposed method is quite tricky to be used in a large construction site with many machines. Theoretically, given a 4 neighborhood movement model and an undirected graph $ G(V, E) $, the branching factor should be estimated as  $(E/V)^k$ and the search space is $V^k$ if k machines should be planned. For 20 agents, considering that a normal working site with $500m \times 100m$ where $50 \times 10$ cells are needed, the search space goes to $ 9.54 \times 10^{53} $, which is unsolvable within acceptable duration for a real application even if the top CPU in 2020 achieving approaching 200 GFLOPs is used. Thus, using the traditional methods to solve the cooperation task is still challenging. Also, some algorithms are based on replanning the path if agents encounter a head-to-head position. However, these methods require excellent perception capability and limit the movement velocity of agents. 

On the other hand, CBS based solution treats all the objects equally. Concretely, each robotics has the same capability, the cell on the ground is assumed as equally challenging to be overcome. However, it does not hold true in a working site; some machines should be assigned a higher priority, and some paths are much easier to pass. For instance, larger machines are usually more challenging to control their velocity. Also, stopping on a slope is much dangerous than on the flat ground. Hence, in our study, we further develop the original CBS so that it can deal with plenty of priority problems in a real working site. Also, the computational time should be much shorter to handle emergence.

\section{Model building}

\begin{figure}
    \centering
    \includegraphics[width=3.5in]{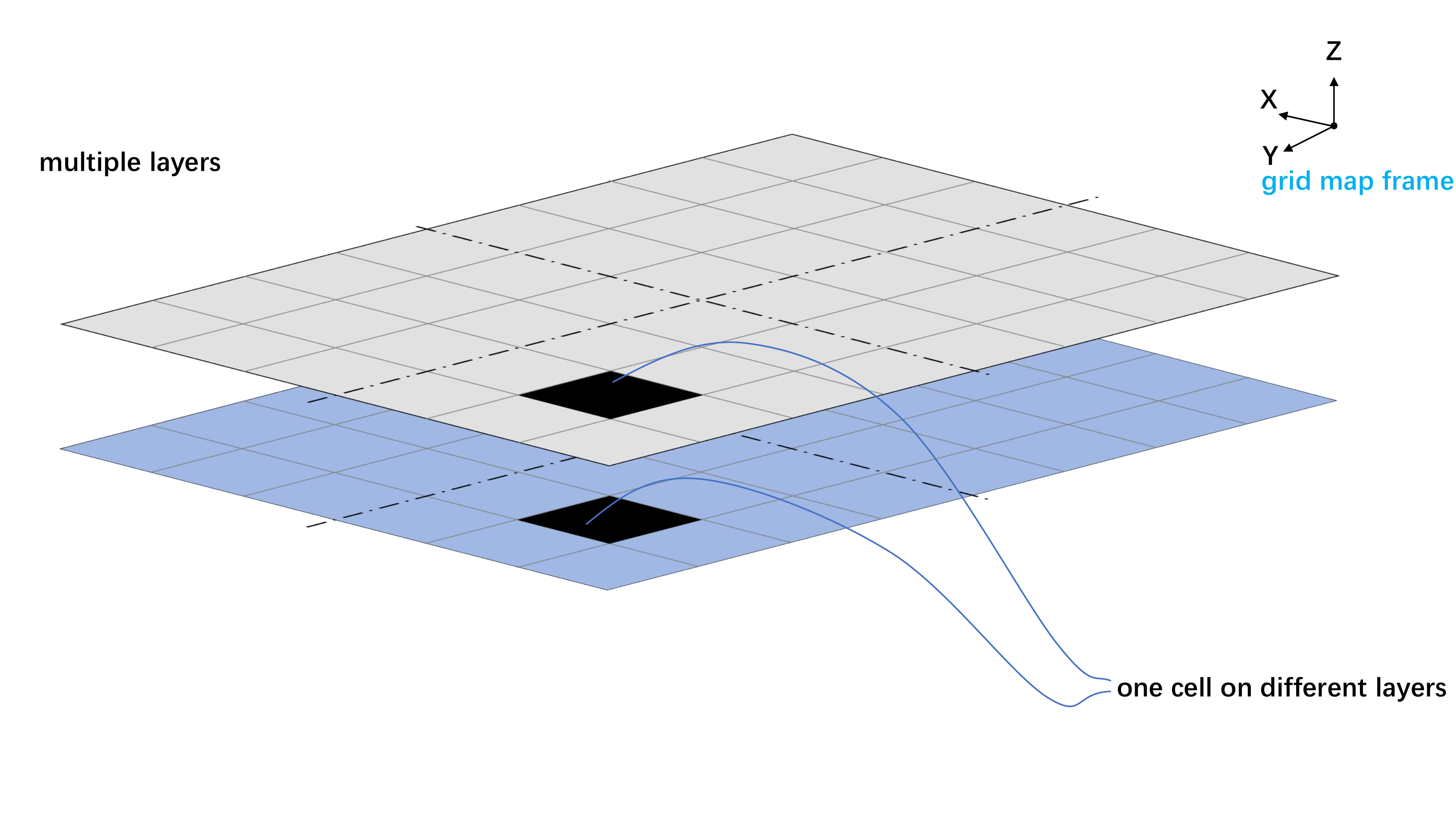}
    \caption{An example of Multilayered grid maps. Our approach depends on multilayered grid maps to offer data of different types of information to make the best path. Concretely, every grid saves a 1*3 matrix, including location information and the corresponding terrain information. The map, which can be visualized in the BIM system, is saved as a  2*m*n*3 tensor, where m is the max displacement in the x-direction, while n is the max displacement in the y-direction. In case a place is unknown, it will be marked as NaN to denote the uncertainty of the regions and be treated as obstacles.}
    \label{fig:A grid-based map uses multilayered grid maps to store data for different types of information}
\end{figure}

\begin{figure}
    \centering
    \includegraphics[width=3.5in]{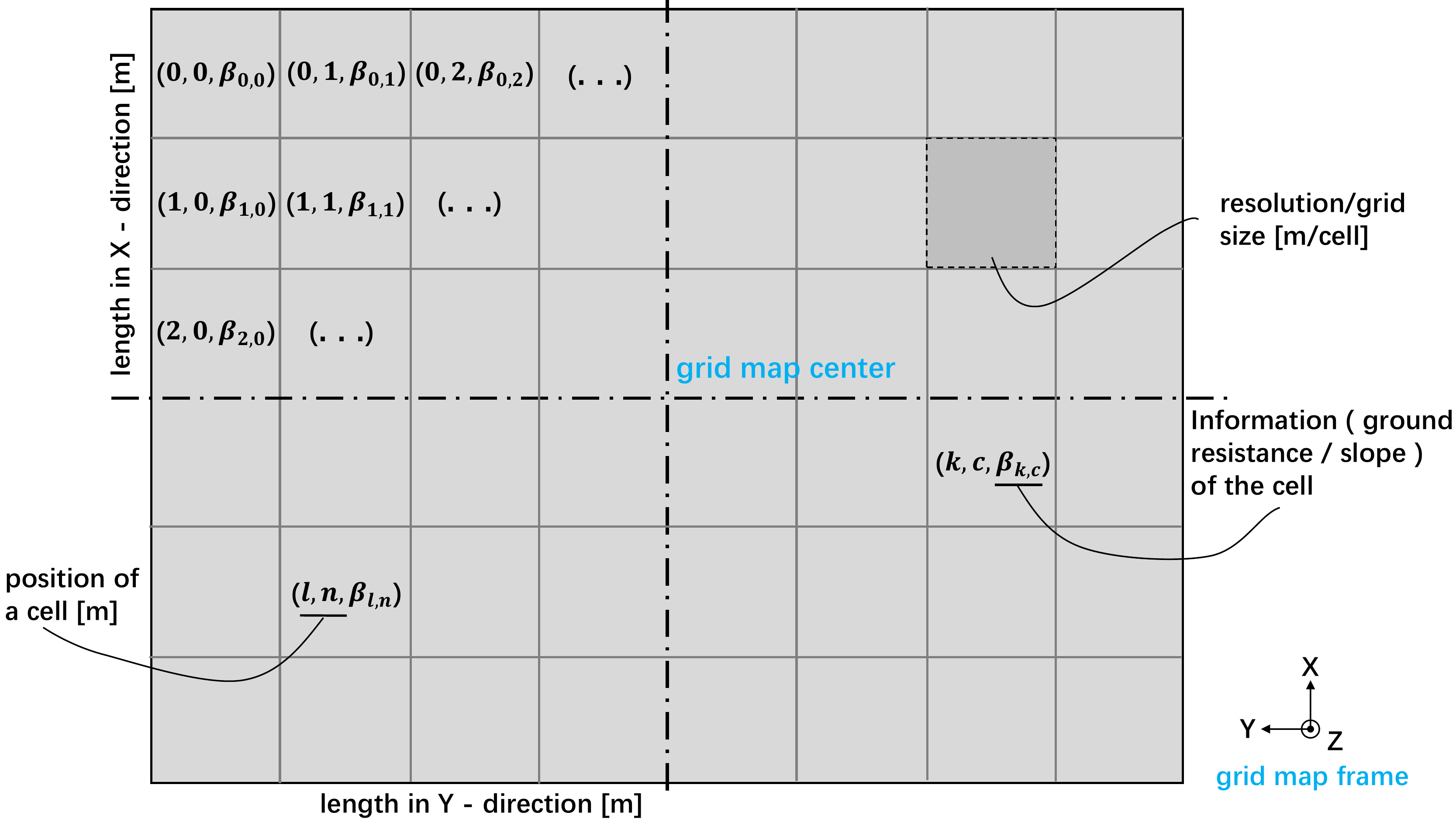}
    \caption{Detail description of a layer in the grid-based map}
    \label{fig:One-layer grid-based map}
\end{figure}

\Figure[h!](topskip=0pt, botskip=0pt, midskip=0pt)[width=6in]{Figures/PDF/construction_site_1.pdf}
{A grid map with terrain weight based on a real construction site. In this map, the green, orange, and red grids demonstrate the easy, normal, and difficult to pass terrain, separately. The blue cells denote the place where it is considered impossible to pass, such as occupied by the obstacles. On a real construction site, the obstacles can be the place to store construction materials temporarily. \label{fig:construction1}}

Nowadays, with the development of SLAM regarding visual recognition, IoT, and satellite technology, the uncertainty degree of construction sites is reducing. Thus we consider the construction sites as known in this paper.  In most instances, the MAPF solution can be evaluated twofold. The first one is sum-of-costs which describes the accumulative cost of all the agents. Such costs can be time consumption, fuel consumption, or some other objective goals. The other way to analyze the performance of MAPF solvers is makespan, indicating the maximal time the last goal has been achieved. Obviously, unlike the robotics in warehouses, working machines perform a relatively long period to do their duty after they have arrived. Thus, makespan is not so vital compared to sum-of costs in the field of construction or milling machines. Consequently, we adopt sum-of-cost as our evaluation criterion rather than makespan. 

For working site MAPF problems, we can describe the problem as given a graph, $G(V, E)$, and a set of k agents labeled as $a_{1}$ . . . $a_{k}$. Each agent $a_{i}$ has a start position $s_{i}$ $\in$ V and goal position $z_{i}$ $\in$  $V$. Based on the practical conditions in a working site, we consider vertex conflicts, edge conflicts, and swapping conflicts as unacceptable conflicts, whereas following and cycle conflicts are allowed in our study. 
Formally, the conflicts are described as, 

\begin{equation}
\left\{
             \begin{array}{lr}
             \pi_i[t]=\pi_j[t], &  \\
             \pi_i[t]=\pi_j[t+1]  	\cup  \pi_i[t+1]=\pi_j[t]
             \end{array}
\right.
\end{equation}
where $\pi_i$ and $\pi_j$ are the single-agent path for $a_i$ and $a_j$ at time step $t$, correspondingly. Apparently, edge conflict is a union of vertex conflict.

\subsection{Multi-layer grid map}

Unlike a standard graph problem which adds the weight on the edges, we add the weight directly on the grid. This is mainly for three reasons. First and foremost, the machines usually occupy a relatively large area and thus should not be modeled as a simple vertex and ignore their geometry. Also, most previous studies in the field of construction machines used the grid-based map. To guarantee compatibility, we tend to use a similar solution since no approach obviously outperforms the other one.  Last but not least, if weights are applied on edges, it becomes challenging to penalize the waiting process.

Inspired by the research from Fankhauser \cite{fankhauser.2016}, a map can include many layers to store different types of data information. Obviously, the map information should be saved in BIM system so that the path planning process can be done. In a previous study \cite{Xiang.2020b}, Xiang developed a realtime map plotter of the construction site according to ground condition, offering multi-layer grid-based maps, which divide the environment into uniform cells. Moreover, maps can also be gathered by Lidar or cameras provided depth information installed on a drone or on the ground.

\Figure[ht!](topskip=0pt, botskip=0pt, midskip=0pt)[width=7in]{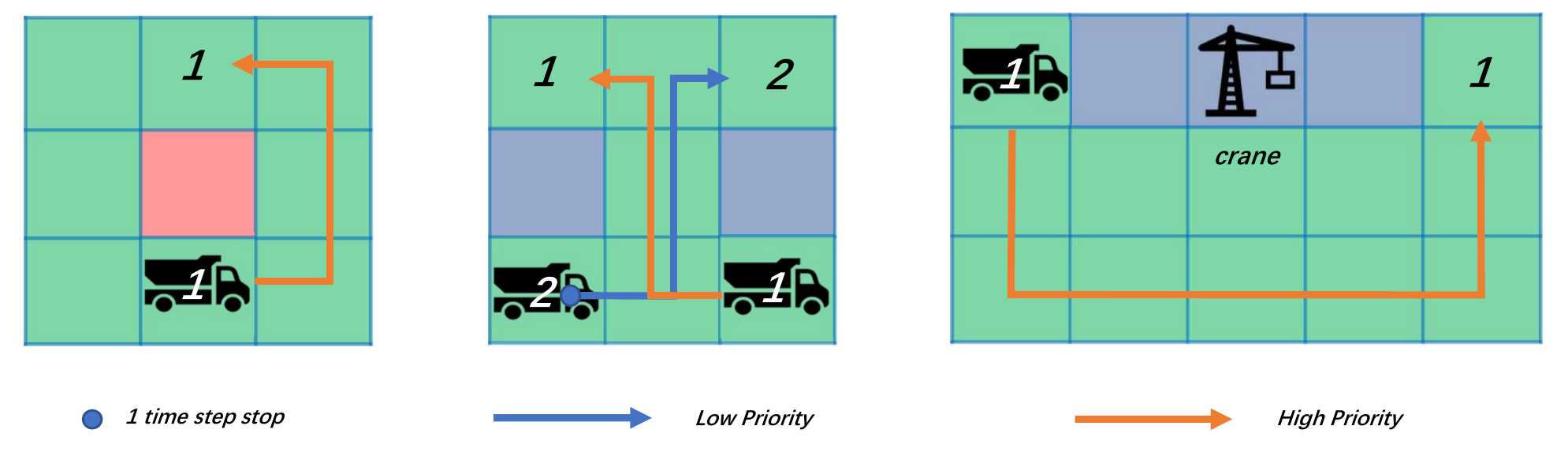}
{The basic requirements of the path planning algorithm. As shown in the left subfigure, the vehicle should take the lowest-cost path to reach its goal. The middle subfigure shows that the vehicle with lower priority should wait until the vehicle with higher priority pass through if there is no other bypass possibility. Last but not least, the right subfigure indicates that a good path should not be too close to dangerous objects. Girds in blue, red, and green represent obstacles, rough road, and normal road separately.  \label{fig:problem}}

Fig. \ref{fig:A grid-based map uses multilayered grid maps to store data for different types of information} illustrates the multilayered grid map concept, where each cell data is stored on the congruent layers. In many construction projects, since resistance and grade of the road are the most of importance information for the construction machines, we show a two layers grid map as an example. Concretely in our study, the map is divided into small cells, whose resolution is 10 meters per cell to cover the geometry of the vehicles. In practice, we can use GPS/IMU 
based Kalman filter algorithms to locate the mobile machine in the construction site. As can be seen in Fig. \ref{fig:A grid-based map uses multilayered grid maps to store data for different types of information}, to describe the ground condition of construction sites, we use the value of each cell to represent the information of the ground situation. In Fig. \ref{fig:One-layer grid-based map}, a layer that holds the data of a grid-based map is shown. Apparently, although we only demonstrate the map with two layers, it is relatively easy to extend the third layer in case more information should be taken into account for the path planning because we can simply add the weights together.

\subsection{Lower level search}

The concept of CBS does not limit the lower-level search algorithm. In addition, since negative weight cannot happen in 3D space, we believe that Dijkstra and $A^*$ can work well for individual shortest path search.  In light of that, we use best-first search. Also, to accelerate our algorithm, we limit the moving direction of an agent to 4 and thus reduce the branching factor to 5, including wait, instead of 9 or more. In order that we can get the optimal solution, we set our heuristics smaller than the real distance since weights are considered. Because grid map is used, we use Manhattan distance as the base of our heuristics to guide our search. 

In the lower level, the algorithm searches the best path of individual agent based on the estimated cost of through current vertex to the goal, formally, 
\begin{equation}
f_{f,r}(n) = \sum\limits_{i=1}^{p} W_{L,i} \cdot g_{i,f,r}(n)+ h_{f,r}(n)
\end{equation}
where $f(n)$ is the estimated cost from its source through current vertex $n$ to its goal, $g_i$ denotes the real cost from the source to the current vertex $ n $ considering the $i_{th}$ weight-grid map, and $h$ denotes the estimated cost from the current vertex $ n $ to the predefined goal based on Manhattan distance. $W_{L}$ is the weight for the specific layer. The index $f$ and $r$ show the estimated cost is forward or backward.

Apparently, planning the best path for machines to reach their goal is a multi-objective task. Generally speaking, the evaluation criterion can be divided into subjective and objective criteria. Obviously, the objective criterion demonstrates the objective criterion of the planned path, especially the terrain which can affect safety and efficiency. As some roads inside the construction site can be built with asphalt, so is considered better road conditions than some road made of sand. Consequently, the cost of passing different routes is different. Besides that, the road slope should be taken into account since waiting on a steep hill is much more dangerous than staying on flat ground. Therefore, we introduce multi-layer to record the individual characteristics of the terrain and plan the best path based on them. Concretely, a construction site map is divided into a series of cells and layers, according to different criteria. The weights of individual cells in one layer are saved as shown in the following matrices.

\begin{equation*}
\Tilde{W}_{m,n} = 
\begin{pmatrix}
w_{1,1} & w_{1,2} & \cdots & w_{1,n} \\
w_{2,1} & w_{2,2} & \cdots & w_{2,n} \\
\vdots  & \vdots  & \ddots & \vdots  \\
w_{m,1} & w_{m,2} & \cdots & w_{m,n} 
\end{pmatrix}
\label{eq:weight_martix}
\end{equation*}

In contrast, the subjective criterion may not harm the whole system's actual performance; however, it has an impact on people's psychology. For instance, a crane or some other dangerous objects, such as a power station, on a working site should be protected, and we should avoid the mobile machines unnecessarily approaching them. As we know, even machines did not involve in an accident, getting close to a dangerous object will be stressful for site managers and indicating a potential risk. To address this problem, we add $g_3$ to penalize the machines for occupying the areas surrounding these special objects. 
\begin{equation}
g_3(n) = \sum\limits_{o=1}^{r} \frac{C_o}{|(X_n-X_o)| + |(Y_n-Y_o)|}
\end{equation}
where $ [X_o, Y_o] $ is the position of the objects which should try to avoid being approached, $C_o$ denotes the intensity.

\begin{algorithm}
    \caption{bidirectional $A^*$ Algorithm at low level to speed up the searching process}
    \begin{algorithmic}[1]
        \renewcommand{\algorithmicrequire}{\textbf{Input:}}
        \renewcommand{\algorithmicensure}{\textbf{Output:}}
        \REQUIRE  $G(v,t), \Tilde{s}, \Tilde{z} $ from predefined map information in yaml, original from visual or Lidar recognition
        \ENSURE  $Path, d_{shortest}$
        \\ \textit{Initialisation} :
        \STATE $ OpenSet  \leftarrow [s, z]$; $ ClosedSet  \leftarrow [\Phi, \Phi]$
        \\$ G^R  \leftarrow ReverseGraph(G) $
        \STATE $dist[:] \leftarrow \infty $, $ dist^R[:] \leftarrow \infty$
        \STATE $dist[s] \leftarrow 0 $, $ dist^R[z] \leftarrow 0$
        \STATE $ [CF, CF^R, proc, proc^R] \leftarrow \Phi $
        \\ \textit{LOOP Process}
        \WHILE{OpenSet[0], OpenSet[1] not $empty$}
            \STATE $ v_c \leftarrow ExtractMin(dist) $, forward otherwise $dist^R$
            \STATE $OpenSet.remove[0](u)$
            \IF{neighbor(u) = valid}  
            \IF{$u$ not in $ ClosedSet[0] $}
            \STATE $ g[0][u]_{tmp}  \leftarrow  g[0][v] + StepCost $ 
            \ENDIF
            \IF{neighbor not in $ OpenSet[0] $}
            \STATE $OpenSet.append(u) $ 
            \ELSIF{ $g[0][u]_{tmp} > g[0][u] $}
            \STATE continue
            \ENDIF
            \STATE $ CF[0][u] \leftarrow v $
            \STATE $ pi_f \leftarrow |\overrightarrow{d(v)-d(s)}|, pi_r \leftarrow |\overrightarrow{d(t)-d(v)}| $
            \STATE $ h_f \leftarrow (pi_f - pi_r) / 2 $
            \STATE $ h_r \leftarrow -p_f$
            \STATE $ f[0][u] \leftarrow g[0][u] + h_f$, if forward otherwise $h_r$
            \ENDIF
            \IF{$ u$ in $  ClosedSet[0] $}
            \STATE break
            \ENDIF
            \STATE $ClosedSet[0].append(u) $
            \STATE repeat symmetrically for $v^R$ as for $v$, where $OpenSet[1]$ and      $ ClosedSet[1]$ should be used
        \ENDWHILE 
        \STATE $distance \leftarrow \infty$,  $u_{best} \leftarrow None $
        \FOR{$u$ in $ClosedSet[0]$ +  $ClosedSet[1]$}
        \IF{$ dist + dist^R < distance $} 
        \STATE $u_{best}   \leftarrow u$
        \STATE $distance   \leftarrow  dist[u] + dist^R[u] $
        \ENDIF
        \ENDFOR
        \STATE $last_0, last_1  \leftarrow u_{best} $
        \WHILE{ $last_0 != s$}
        \STATE path.append(last)
        \STATE $last_0 \leftarrow CF[0][last_0]$
        \STATE path.reverse
        \ENDWHILE
        \WHILE{ $last_1 != t$ }
        \STATE path.append(last)
        \STATE $last_1 \leftarrow CF[1][last_1]$
        \ENDWHILE
        \RETURN $path, distance$ 
    \end{algorithmic} 
\end{algorithm}

To accelerate the low-level searching process, we utilize the bidirectional $A^*$ algorithm with a path return for the initial search. The following Algorithm 1 shows the algorithm's steps, where CF is the dictionary to save the path sequence. Since bidirectional $A^*$ is not suitable for dealing with the waiting operation, neighbors are valided if the grids are not occupied by obstacles, i.e., ignoring the conflicts with other agents. The index 0 in the algorithm denotes forward, whereas index 1 means backward. 

\begin{figure}
    \centering
    \includegraphics[width=3.3in]{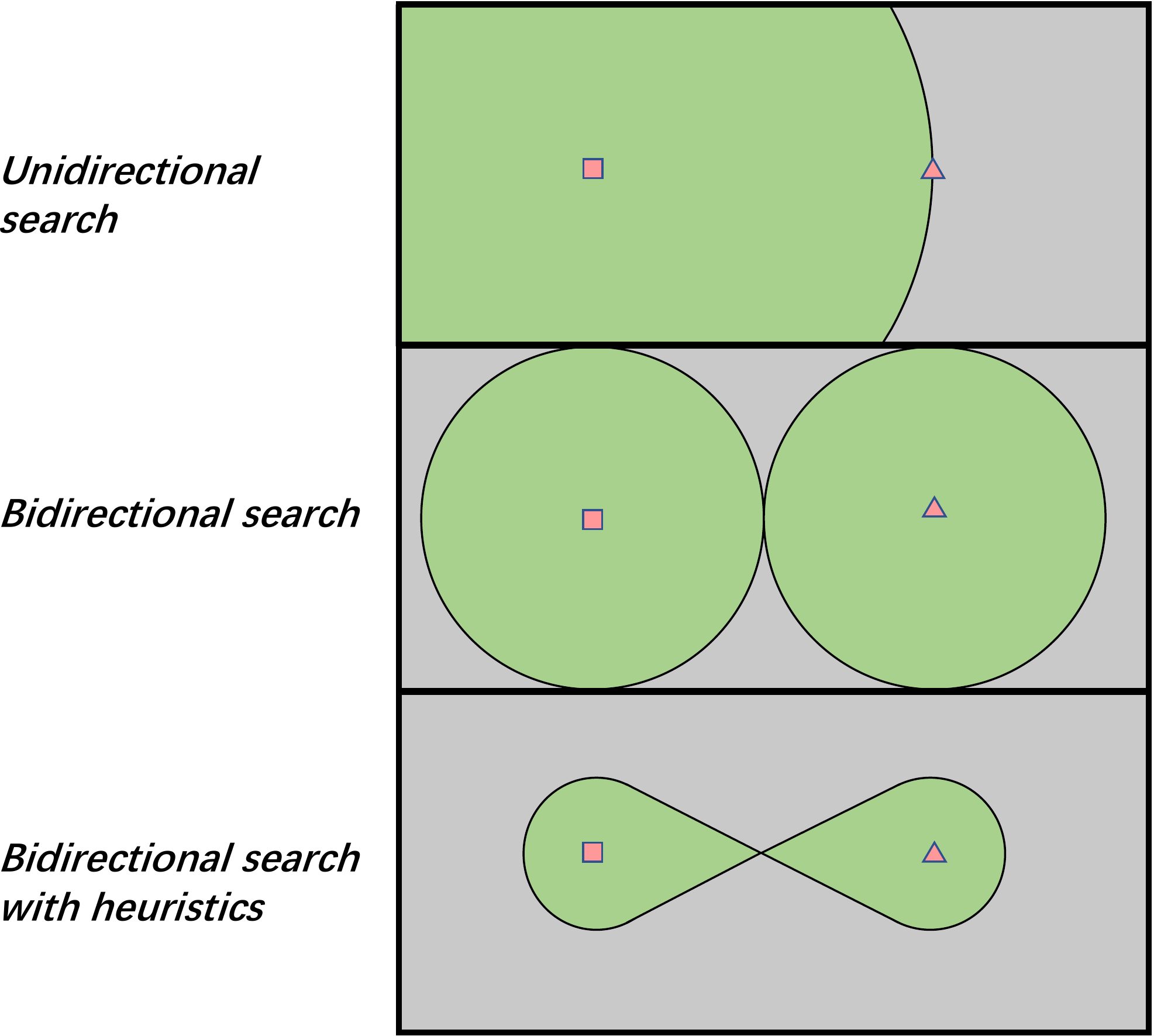}
    \caption{Illustration of the benefit of bidirectional search with heuristics. We use squares and triangles to represent the start point and the goal point separately. Here the grey region denotes the whole region of the map, and the green region shows the region that the algorithm must search before finding the shortest path for agents. Apparently, bidirectional search with heuristics cannot be slower than its antecessors.}
    \label{fig:benefitBidAstar}
\end{figure}

After the initial solution is proposed, we adopt unidirectional $A^*$ to solve the conflicts among the agents. Also, the priority of the agents is taken into account here. As a conflict occurs, the algorithm will give the order which agents should avoid other higher priority agents. Unlike some other research using hard priority, which might lead to the algorithm become not complete, we adopt the soft priority, namely penalization function, to guarantee the algorithm to find the feasible solution if the problem is a solvable MAPF problem. Since the $A^*$ algorithm is well known, we only give the part where involves in the priority of the agents, in Eq. \eqref{eq:uniAstar},

\begin{equation}
g[u]_{tmp} = g[v] +  P_{a_i} \cdot \sum\limits_{k=1}^{L} StepCost 
\label{eq:uniAstar}
\end{equation}
where u is the neighbour point and v is the current point. L is the total layers of the map, and $P_{a_i}$ is the priority value of the agent $i$.

\subsection{Higher level search}

Although the papers about CBS and its variants showed the success rate of the algorithms under various specific scenarios, they are performed with a time limit of at least 1 minute. Considering that some machines might not maintain their speed and emergencies may happen, we believe a feasible algorithm for the construction site should give a command to all the participants within 5 seconds even the order can be just wait. 

\begin{algorithm}
    \caption{High level realtime search}
    \begin{algorithmic}[1]
        \renewcommand{\algorithmicrequire}{\textbf{Input:}}
        \renewcommand{\algorithmicensure}{\textbf{Output:}}
        \REQUIRE $G(v,t), \Tilde{s}, \Tilde{z} $ from predefined map information in yaml, original from visual or Lidar recognition
        \ENSURE  solution for all agents, total cost
        \\ \textit{Initialisation} :
        \STATE $ start.constrains \leftarrow \Phi$
        \STATE $ start.solution, start.cost \leftarrow bid\_Astar.search()$ 
        \STATE $ allConflicts \leftarrow findConflicts(start.solution)$
        \STATE insert start to OPEN
        \\ \textit{LOOP Process}
        \WHILE{OPEN not Empty}
            \STATE $P \leftarrow$ the node with lowest solution cost
            \IF {$ counter >  threshold $}
                \STATE $P.remove(allConflicts)$   (remove some agents)
            \ENDIF
            \IF {Validate($P$) = 1 (no conflicts found)}
                \STATE return $P.solution, P.cost$
            \ENDIF
            \STATE $C \leftarrow$  first conflict $(a_i, a_j, v, t)$ in $P$
            \STATE $ allConflicts.append(C), counter++ $
            \FOR{$a_c$ in $C$}
                \STATE $ND \leftarrow P $
                \STATE $ND.constraints.append(a_c, v, t) $
                \STATE $ND.solution.update(uni\_Astar.search()) $
                \STATE $ND.cost.update( SIC(ND.solution)) $
                \STATE Insert $ND$ to OPEN
            \ENDFOR
        \ENDWHILE
    \end{algorithmic} 
\end{algorithm}

Concretely, we let some machines have priority to move to their goal while the others should wait for a while or find a midway destination in case when the task is too complicated for a realtime response. Or the algorithm suggests to reduce the total mount of machines on site if necessary. Different from the original CBS algorithm, we add four different strategies in case the algorithm cannot find a solution for all agents within 5 seconds. The basic ideas of these acceleration process are reducing the complexity of the task by solving the problem step by step and described as follow,  
\begin{itemize}
    \item Directly remove the agents causing most conflicts, including initial conflicts and update conflicts, found by line 3 and line 13 in Algorithm 2. Since the computation time of the CBS-based algorithm depends on the conflicts number, reducing the trouble makers can surely speed up the searching process.   
    \item Let the machines have the same moving direction to move first. Obviously, conflicts can be avoided if all the agents move in the same direction.
    \item Randomly select some machines in independent sub-regions to move first.
    \item Let the machines having lower estimated cost move first. 
\end{itemize}

The individual path will be compared at the high-level search to find out the conflicts among the agents. We use bidirectional $A^*$ algorithm with the heuristics proposed by \cite{Ikeda.1994} to create the initial path of each agent in order to enhance the realtime performance, and afterward utilize unidirectional $A^*$ to update the individual path of each agent since only unidirectional $A^*$ can deal with the waiting process. In the BIM system, during the searching process, the algorithm saves the conflicts position and the corresponding agents. As mentioned, in case the algorithm cannot solve the planning problem due to too many conflicts, the algorithm will remove some agents and then replan the paths of the rest agents. In this fashion, we ensure that the known MAPF problems can be solved in a timely manner. If emergence occurs and the algorithm cannot solve the new task in time, we can easily and quickly locate the trouble maker. Apparently, the optimization direction here is not only to ensure a short calculation time, but also to make as many machines as possible move at the same time.

\Figure[t!](topskip=0pt, botskip=0pt, midskip=0pt)[width=3.3in]{Figures/PDF/construction_site_3.pdf}
{Map example. Another map based on a real construction site which has more narrow corridor. \label{fig:construction2}}

\section{Experiment on real working sites}

\Figure[ht!](topskip=0pt, botskip=0pt, midskip=0pt)[width=7in]{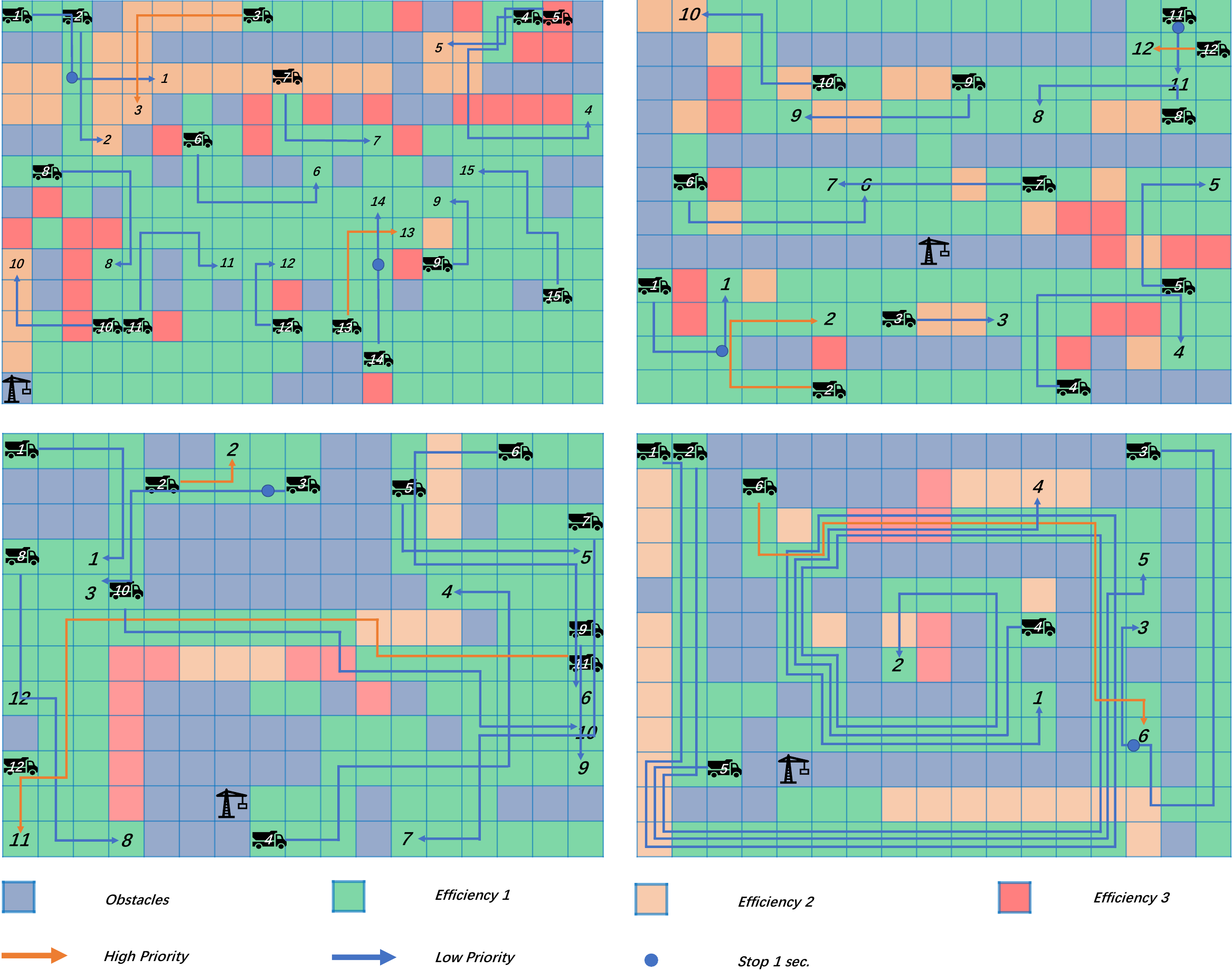}
{The planned path for each map. In the clockwise direction, the subfigures demonstrate the final solutions, including the best path for each machine for maps 1, 3, 4, 5. The left bottom point is defined as the original point (0,0), and the horizontal axis is the first axis.  The layout of map 2 is the same as map 1; however, the difference is that there are more agents on map 2. Due to its huge amount of information, we give the planned schedule in Tab. \ref{tab:scheduleMap2} instead of using figures.   \label{fig:PathSolution}}

As a consensus of the research in the field of graph theory, although a conclusion about a specific map cannot guarantee its effectiveness on another map, the closer the map, the closer the effect. In order to show the benefit of the introduction of MAPF in the mobile machines industry, we validated our algorithm on five typical real working sites. Concretely, they are a relatively open field with 20 or 50 agents, an open field with many obstacles, a two-side working site connected by a bridge or narrow corridor, and a typical mining site. Since the map will also be shown in the path results as background, here we only demonstrate how a map will be processed to give the prior information for successfully pathfinding on the first map and third map to avoid repetition. The maps shown in Fig. \ref{fig:construction1} and Fig. \ref{fig:construction2} are on the same site at different times. Obviously, Fig. \ref{fig:construction1} is the earlier stage while Fig. \ref{fig:construction2} shows the later stage as the construction process proceed since more facilities are there. In our experiments, the dimensions of our maps are  $20 \times 13 $ and $17 \times 12 $. For the sake of simplicity, we also assume the velocity of all the machines are constant; however, it is surely easy to achieve the situation that the machines have quite different speed since we can use the fastest speed as a reference and allow the slower agents occupy more than one grid at the same time or vice versa. 

As we can imagine, the faster the project, the faster the construction site changes. This indicates the difficulties of using a pre-calculated path planning for a construction site.  In this study, we divided the grids into different regions with respect to whether the road is easy to be passed through, the slope of the road, and whether the place is safe. 

\begin{table}[!ht]
	\caption{Weight table. The weights we use to describe the complicated terrain of construction sites.}
	\centering
	\begin{tabular}{c|c}
	\hline \hline
      Layers & Weight  \\ \hline
        Roughness & [1, 5, 9]  \\ \hline
        Slope & [1, 5]  \\ \hline
        Safety & [1, 15]  \\ 
	\hline 
	\end{tabular}
	\label{tab:weightParameters}
\end{table}

Here we show the solution finding time on CPU core i7 4720HQ@ 2.6GHz. Because of its reasonable price at the end of 2020, it is suitable for large-scale commercial use. To reduce the randomness, we did the experiments 50 times and gave the average finding time, and the average number of conflicts occur to analyze the conflicts and thus show the rationality of our optimization. Notice that we rounded the numbers to one decimal place if any. 

\begin{figure*}[!t]
        \newcommand{\w}{0.42}
        \centering 
        \subfloat[][Vertex conflict position]{
            \includegraphics[width=\w\textwidth]{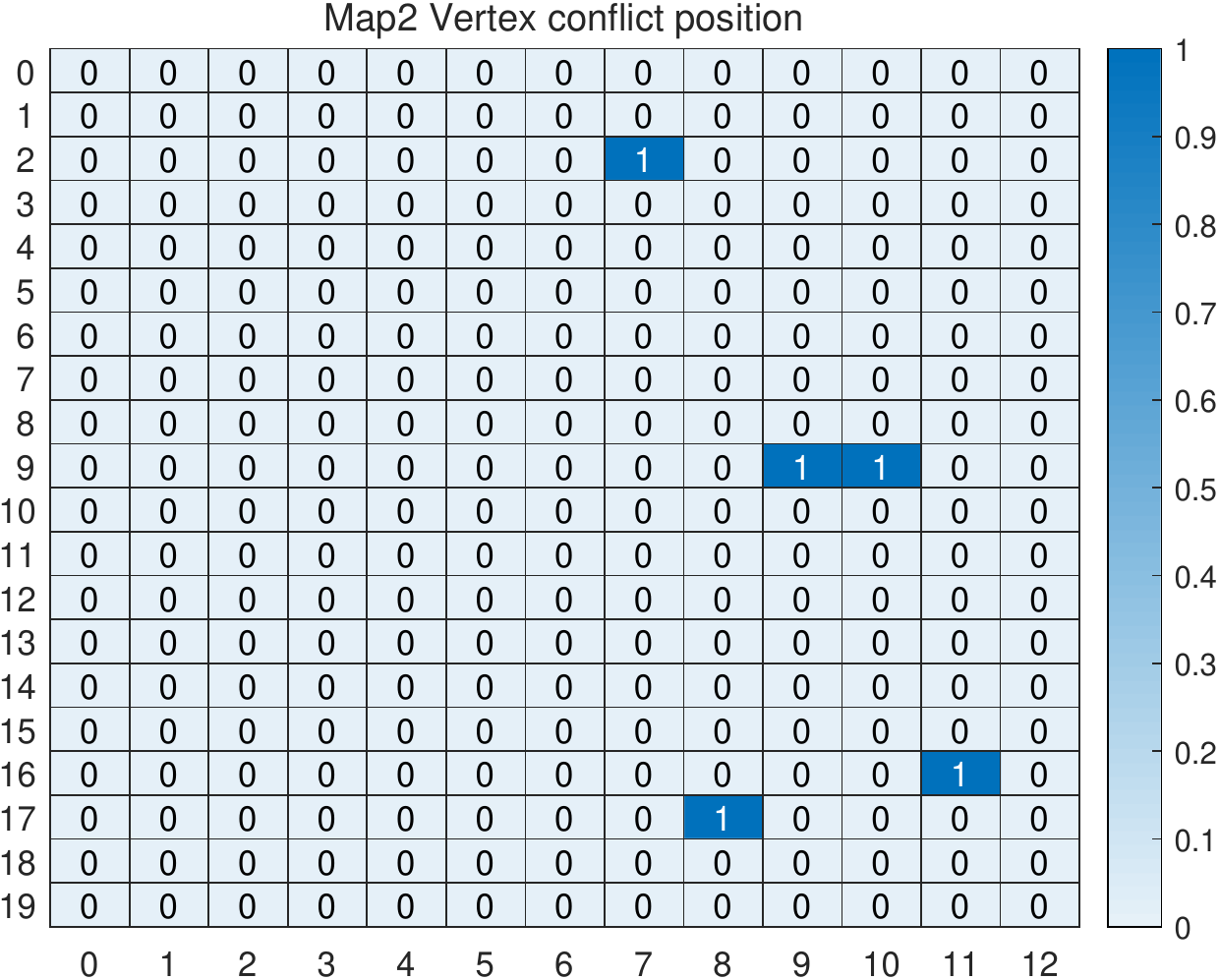}}
            \hfil 
        \subfloat[][Updated vertex conflict position]{
            \includegraphics[width=\w\textwidth]{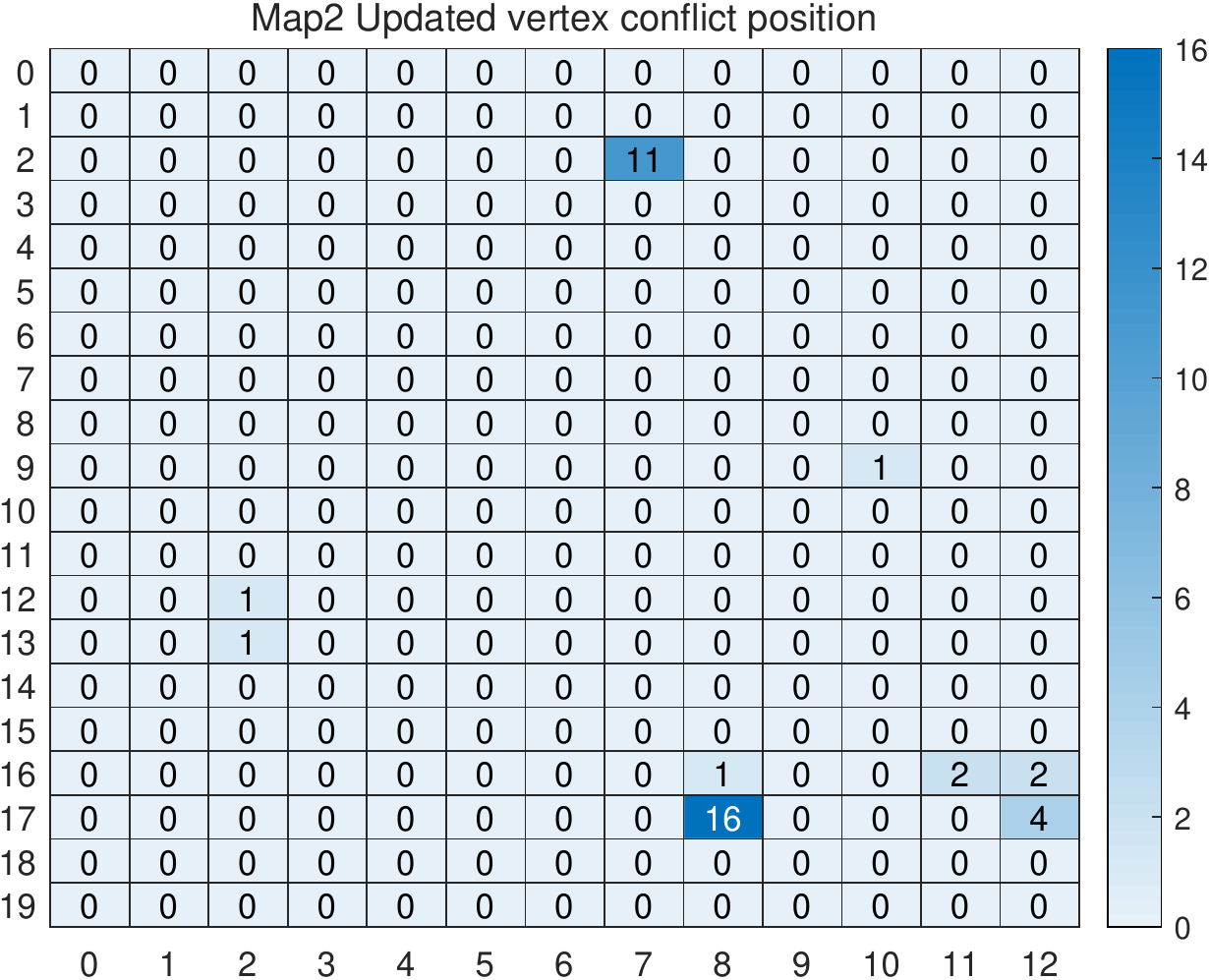}}
            \hfil 
        \subfloat[][ \centering Edge conflict position]{
            \includegraphics[width=\w\textwidth]{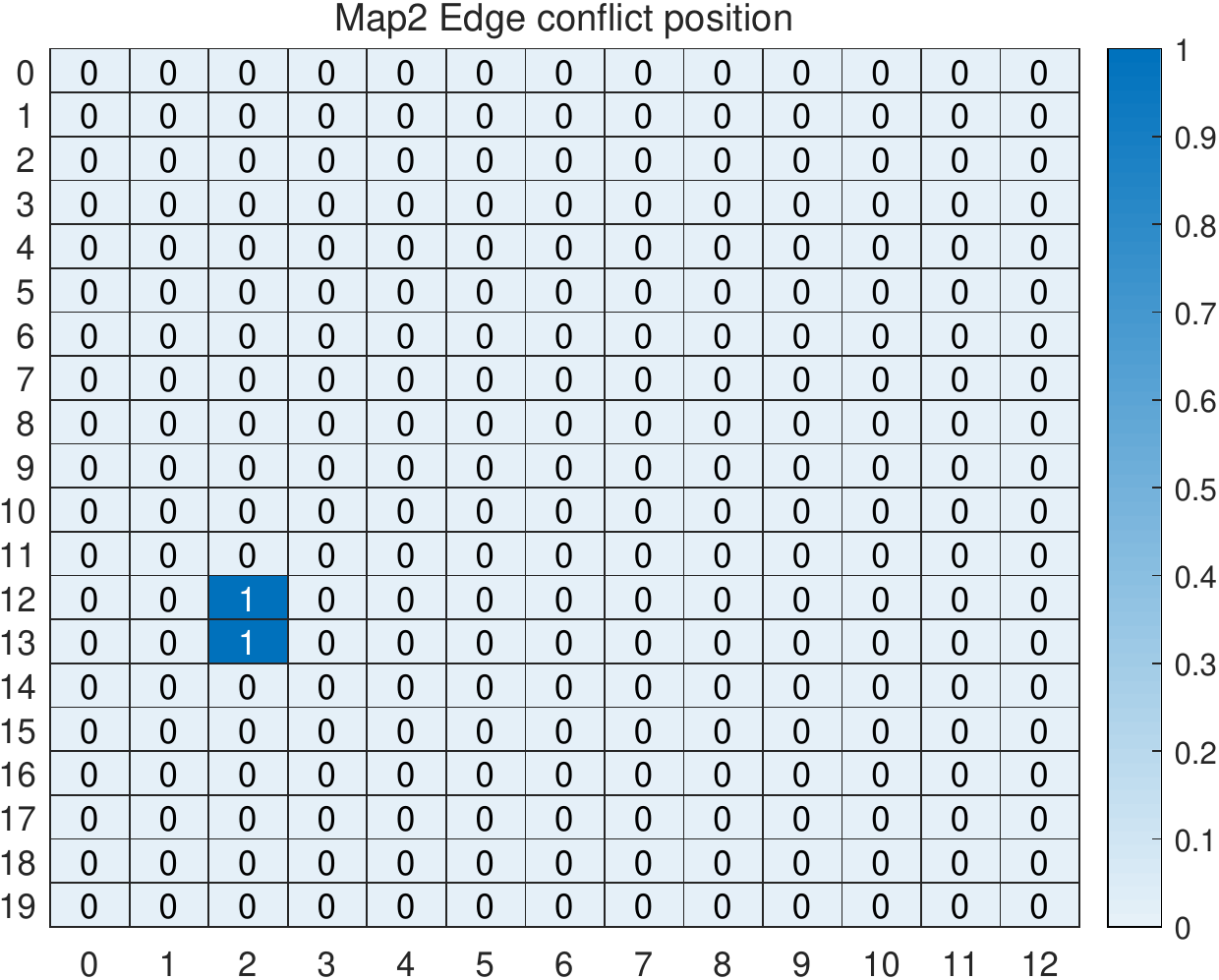}} 
            \hfil 
        \subfloat[][\centering Updated edge conflict position]{
            \includegraphics[width=\w\textwidth]{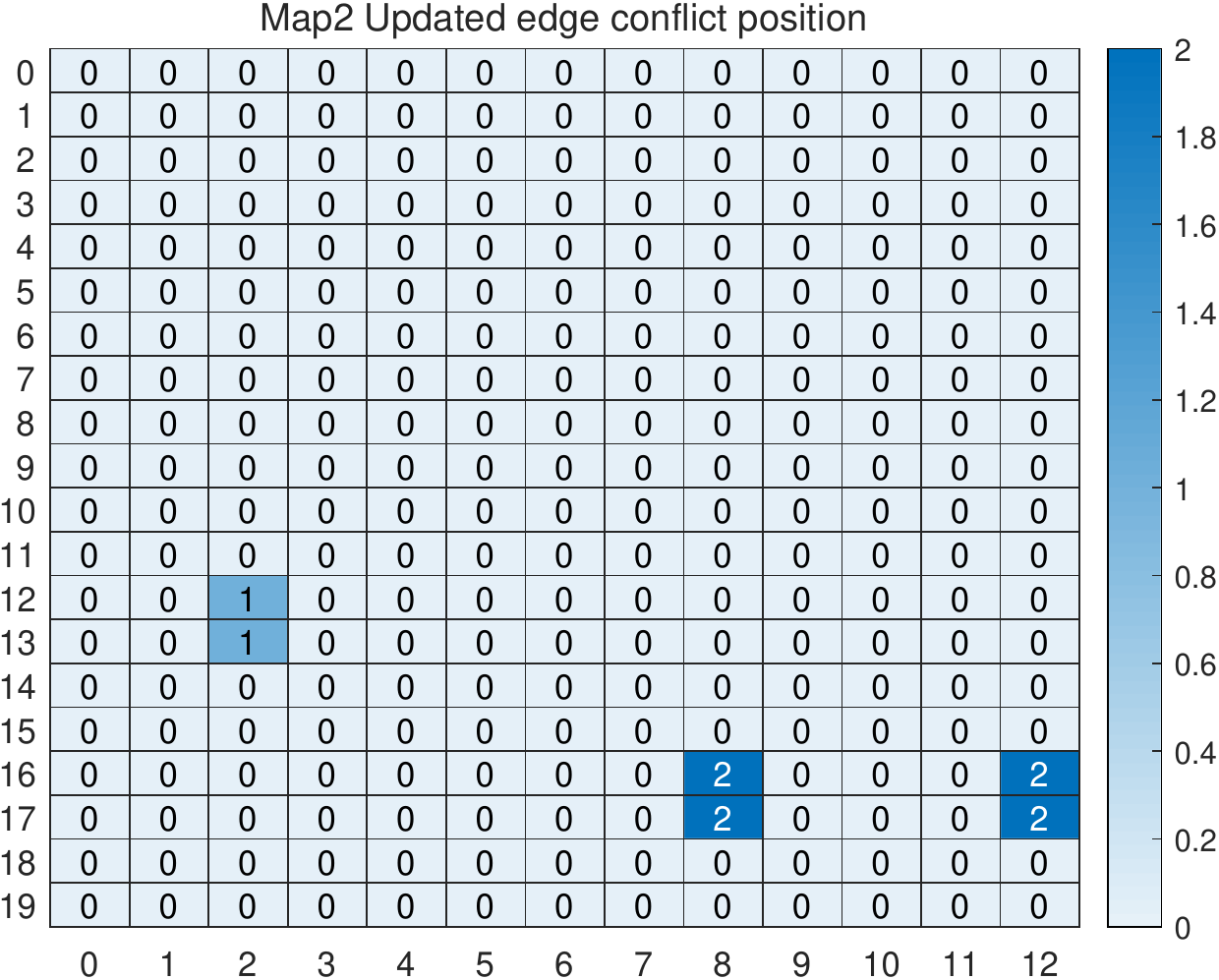}} 
        \caption{The place where the agents intend to pass through and the resulting conflicts on the second map. As we can see from the points allocation, initial conflicts have a great relevance to the conflicts occurring during the conflicts avoidance process. Notice that we did the experiment 50 times and the conflicts shown in these figures are the average number of these 50 experiments.} \label{fig:conflictsPositionMap2}
\end{figure*}

Before we analyze the results of our experiments, we summarize the basic ideas of the algorithm we used. Similar to the original CBS algorithm, our MAPF algorithm also adopts a two-level search, where the upper level finds the conflicts among the agents and the lower level search the best path for individual agents. The lower level finds the path first and sends the initial proposal to the upper level. Afterward, the upper level will check whether the planned path has a or many conflicts with others. In case there are no conflicts, the center commander, an AI system, agrees to the preliminary proposal to become the final solution of MAPF and all agents are allowed to execute this solution. In other cases, if there have some conflicts, the upper level will find out the conflicts and send this information as constraints to the lower level to avoid the conflicts. To generate the initial individual path for each machine faster, we use bidirectional search. And then update the individual path if the upper level finds out a conflict with unidirectional $A^*$ algorithm. The algorithm tries to modify the solution having the lowest cost, which guarantees the solution to be optimal. In the experiments, we do not assume what the participants are, nor do we assume its working process to ensure the generalization of our method.

\section{Experiment results}

\begin{figure*}[!t]
        \newcommand{\w}{0.42}
        \centering 
        \subfloat[][Vertex conflict position]{
            \includegraphics[width=\w\textwidth]{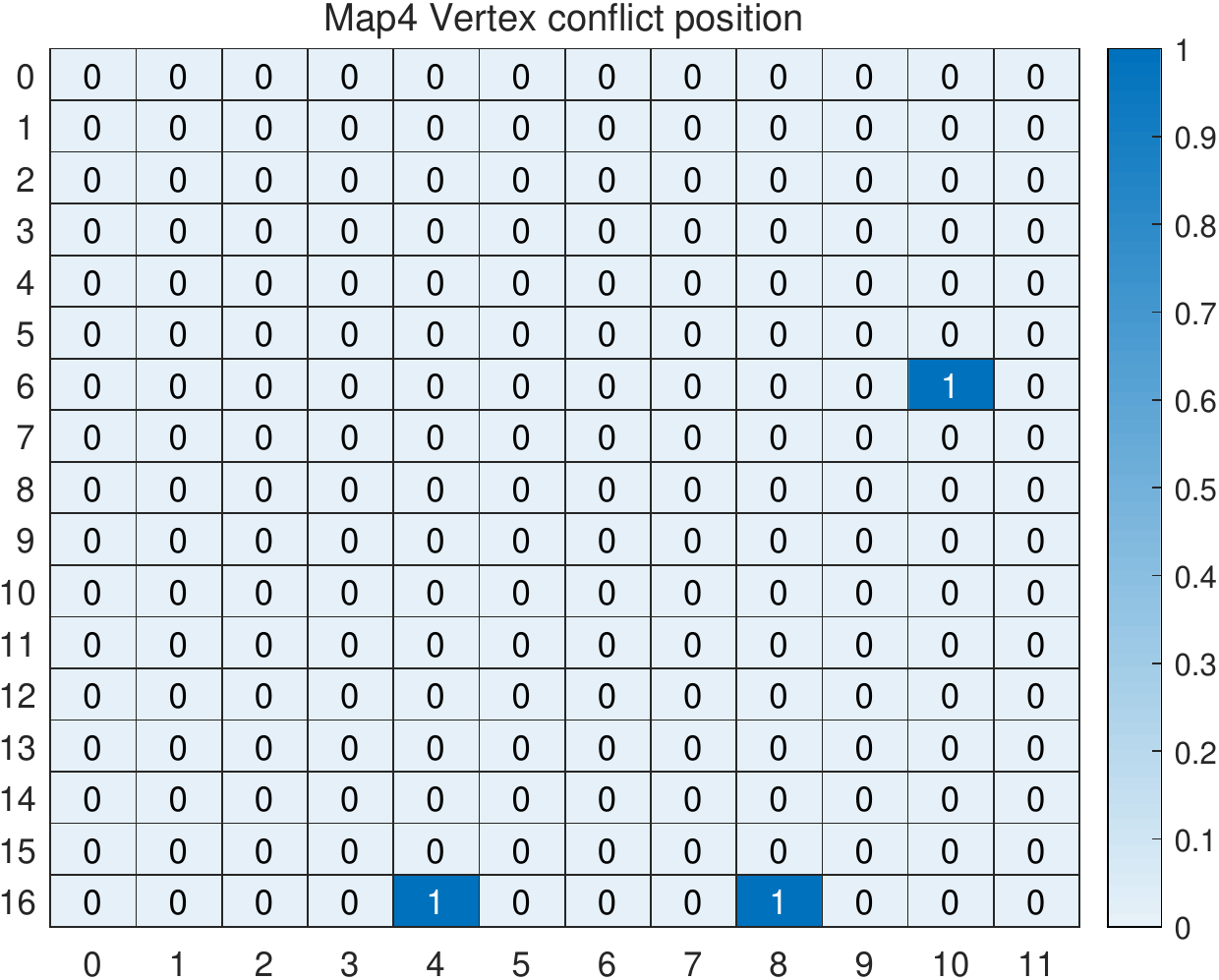}}
            \hfil 
        \subfloat[][Updated vertex conflict position]{
            \includegraphics[width=\w\textwidth]{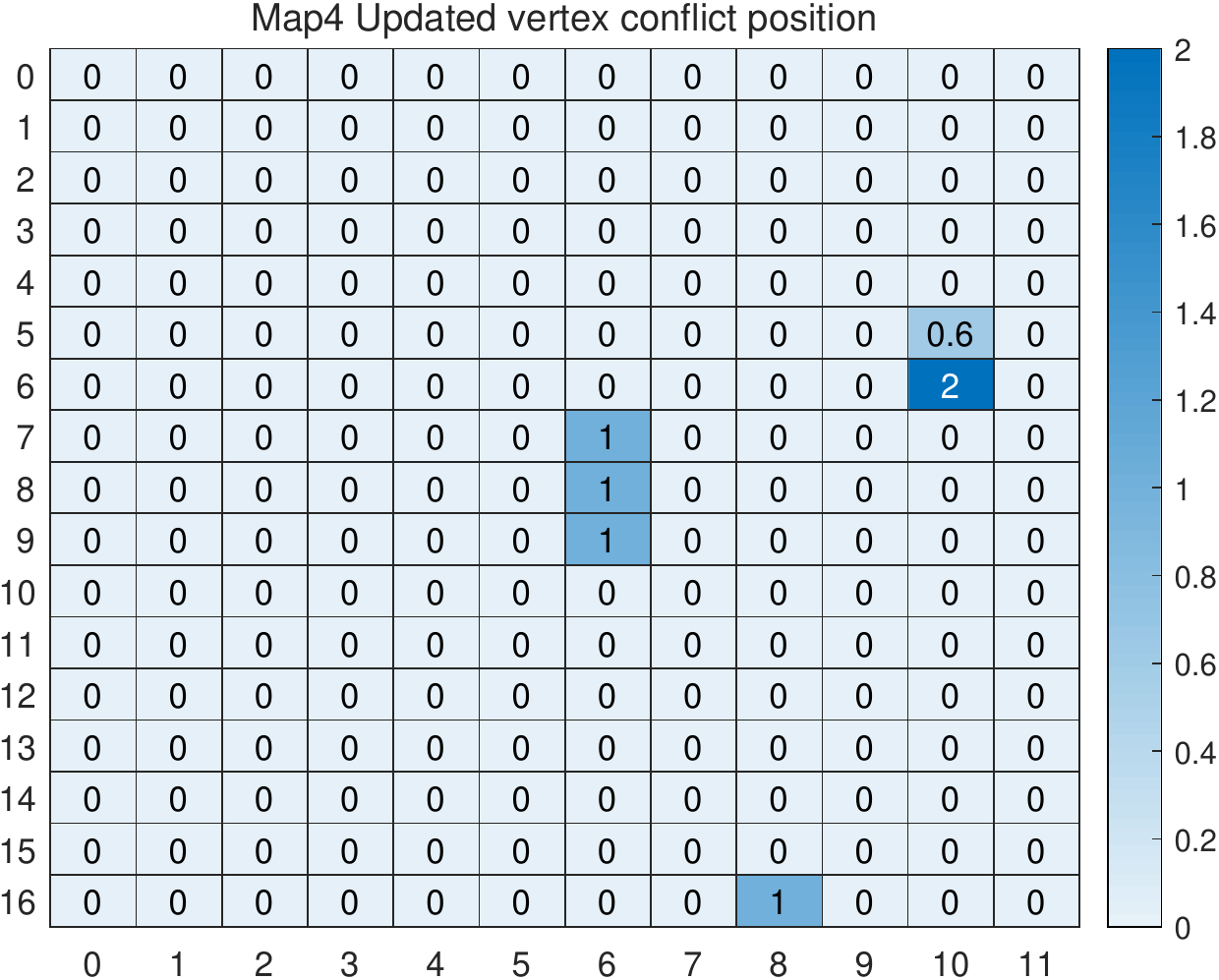}}
            \hfil 
        \subfloat[][ \centering Edge conflict position]{
            \includegraphics[width=\w\textwidth]{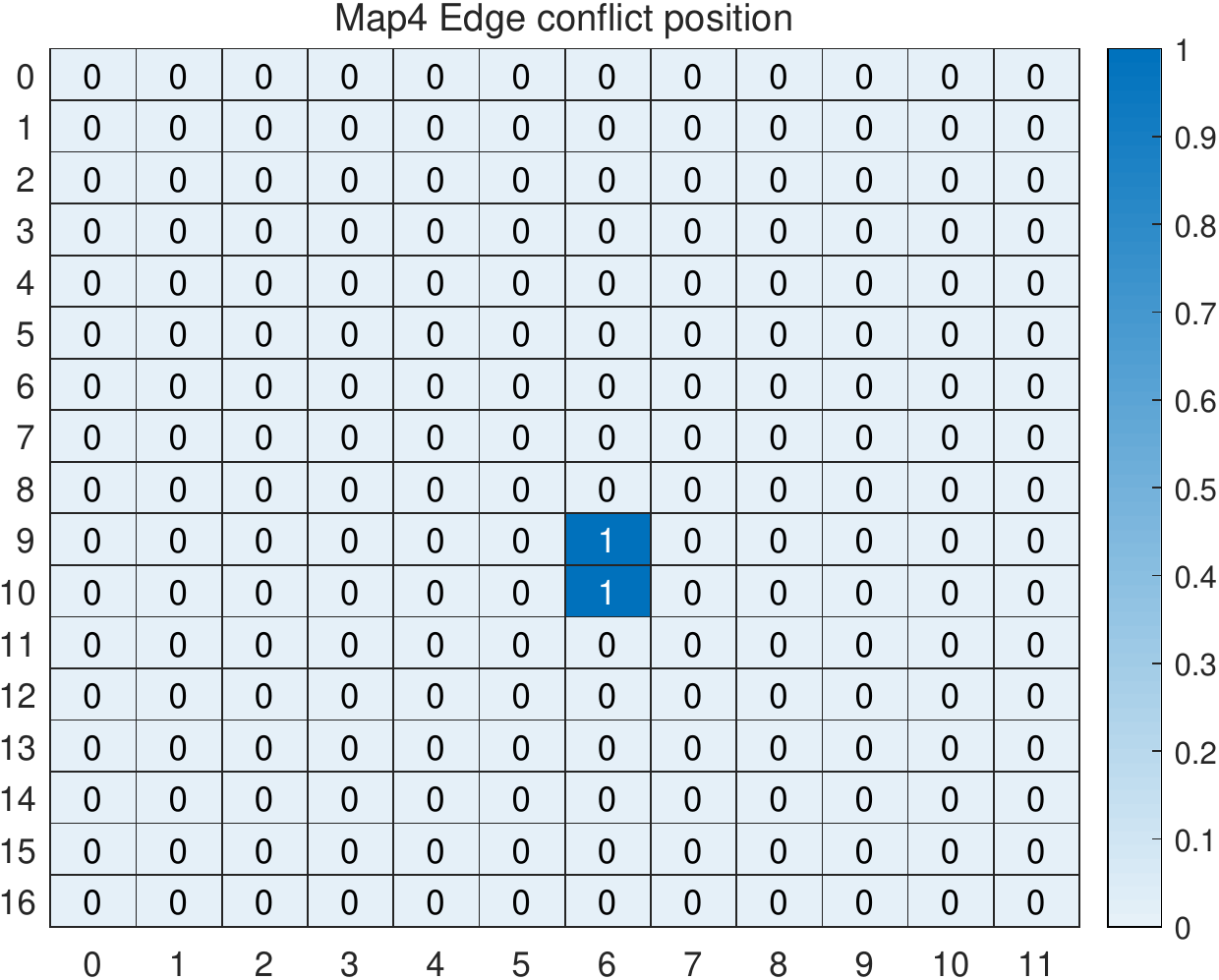}} 
            \hfil 
        \subfloat[][\centering Updated edge conflict  position]{
            \includegraphics[width=\w\textwidth]{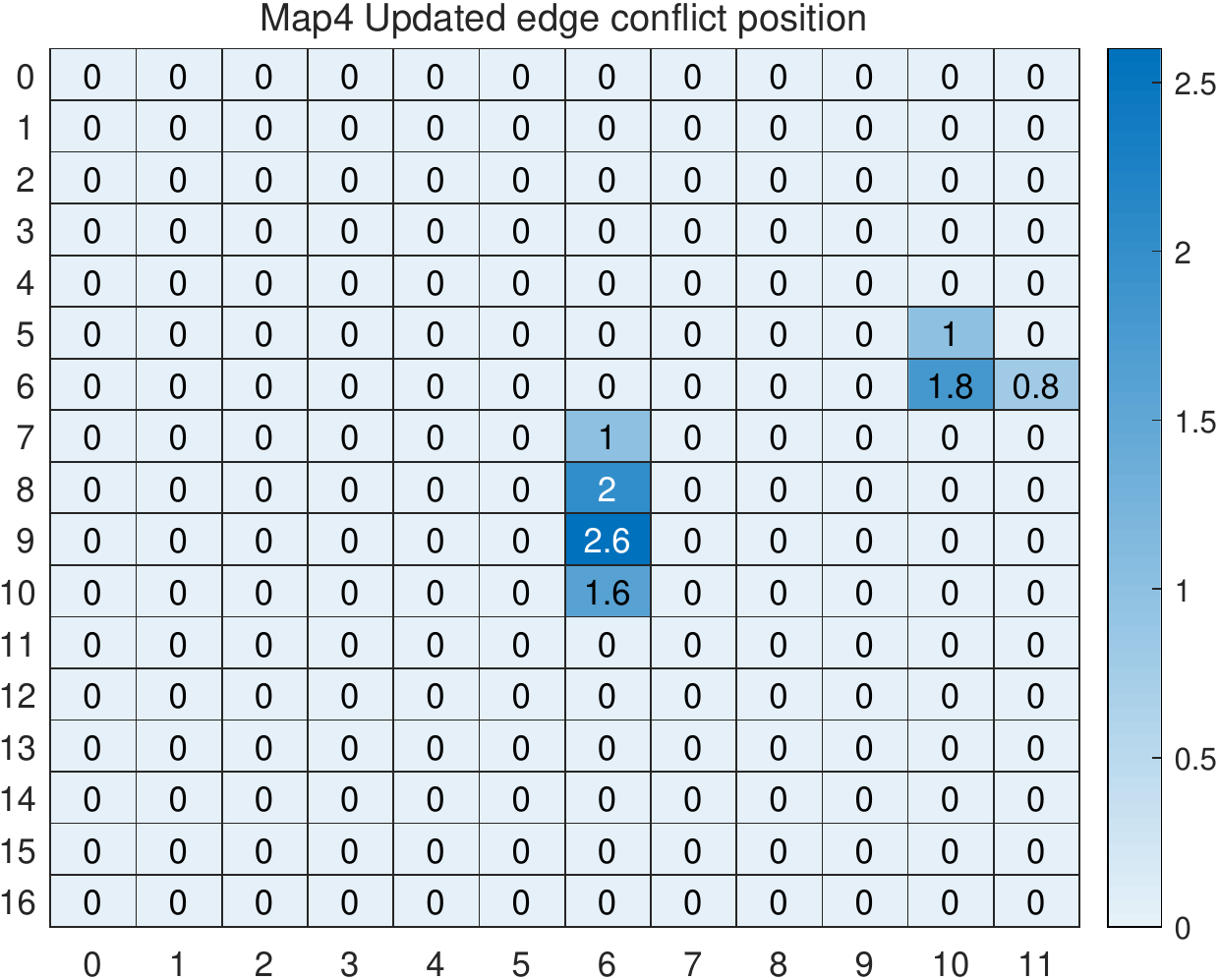}} 
        \caption{The place where the agents intend to pass through and the resulting conflicts on the fourth map. As we can see from the points allocation, initial conflicts still have a great relevance to the conflicts occurs during the conflicts update process.} \label{fig:conflictsPositionMap4}
\end{figure*}

\begin{figure*}[!t]
        \newcommand{\w}{0.42}
        \centering 
        \subfloat[][Edge conflicts and updated edge conflicts]{
            \includegraphics[width=\w\textwidth]{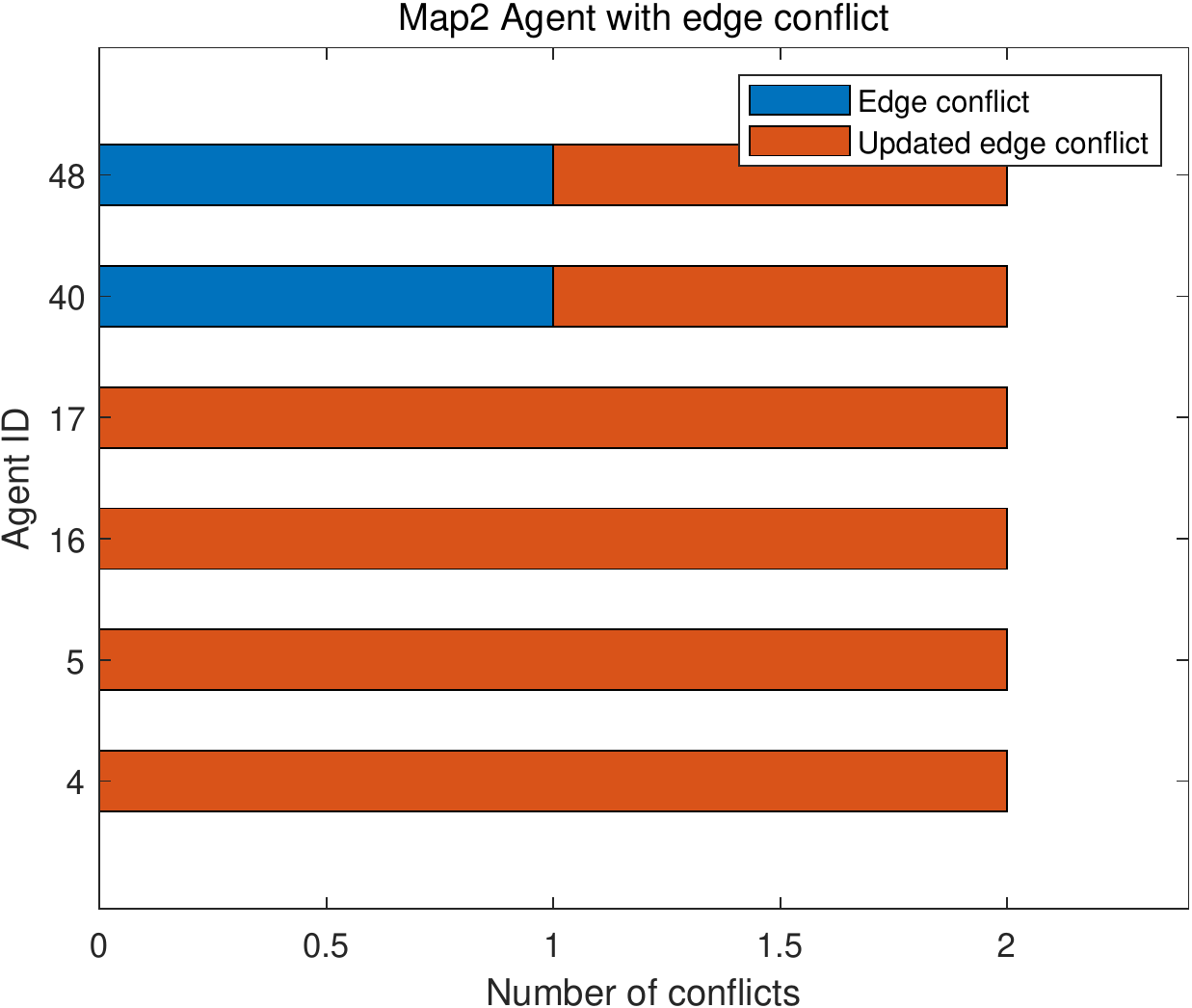}}
            \hfil 
        \subfloat[][Vertex conflicts and updated vertex conflicts]{
            \includegraphics[width=\w\textwidth]{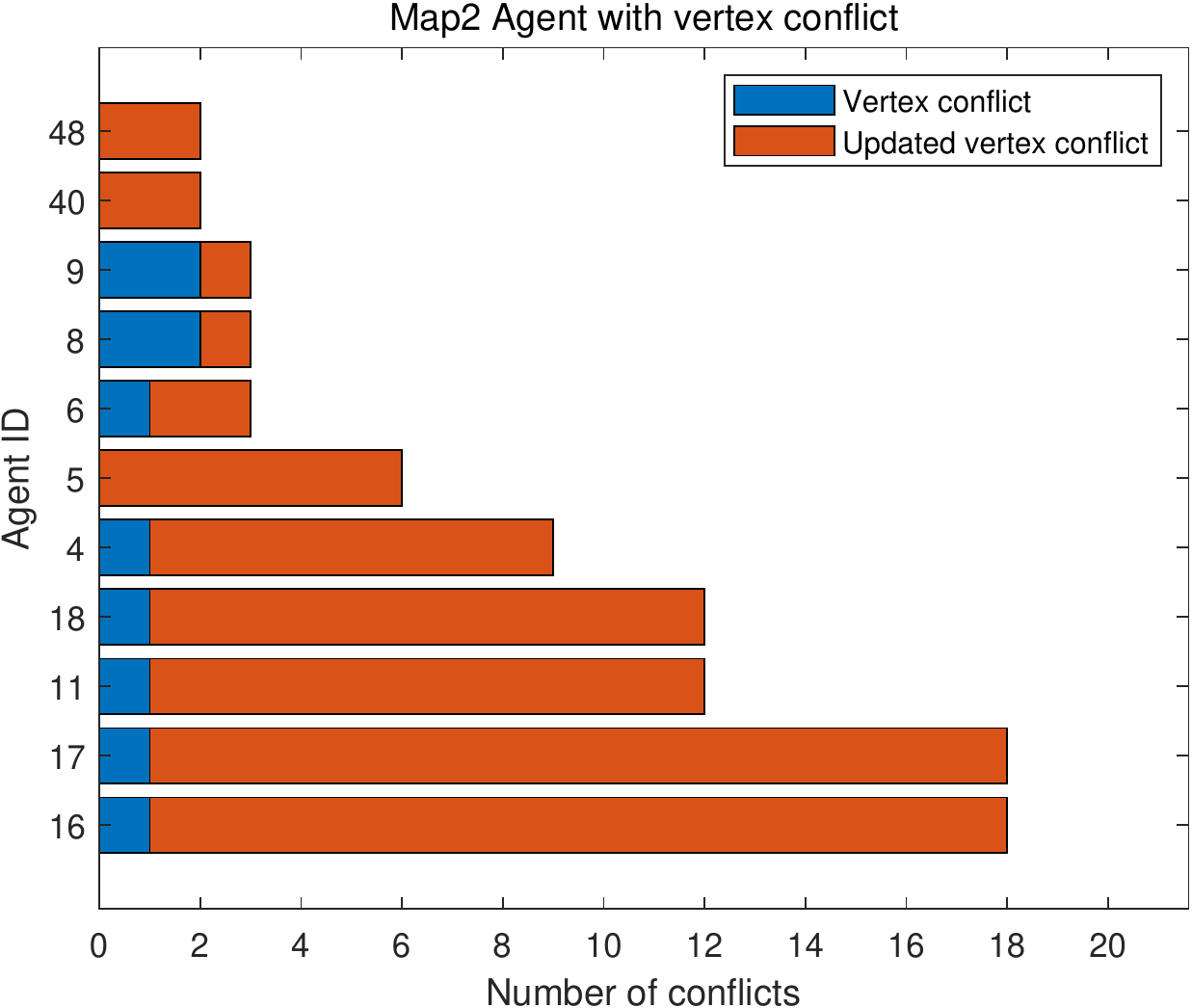}}
            \hfil 
        \caption{Statistics of the conflicts made by corresponding agents on map 2.  Here blue histogram denotes the conflicts found by initial bidirectional search, and the orange histogram shows the conflicts solved while updating the solution with unidirectional search.} \label{fig:TroubleAgentsMap2}
\end{figure*}

\begin{figure*}[!t]
        \newcommand{\w}{0.45}
        \centering 
        \subfloat[][Edge conflicts and updated edge conflicts]{
            \includegraphics[width=\w\textwidth]{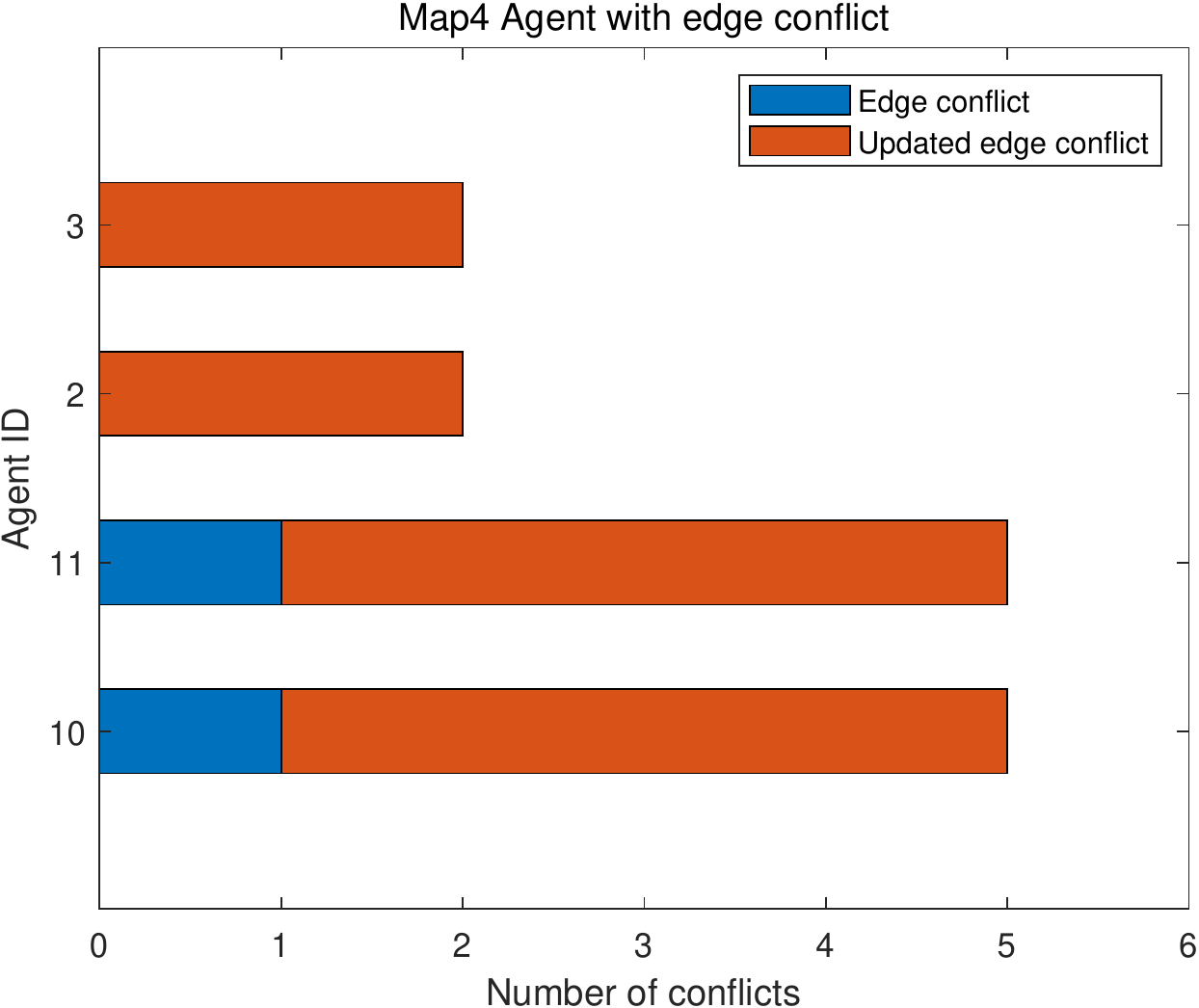}}
            \hfil 
        \subfloat[][Vertex conflicts and updated vertex conflicts]{
            \includegraphics[width=\w\textwidth]{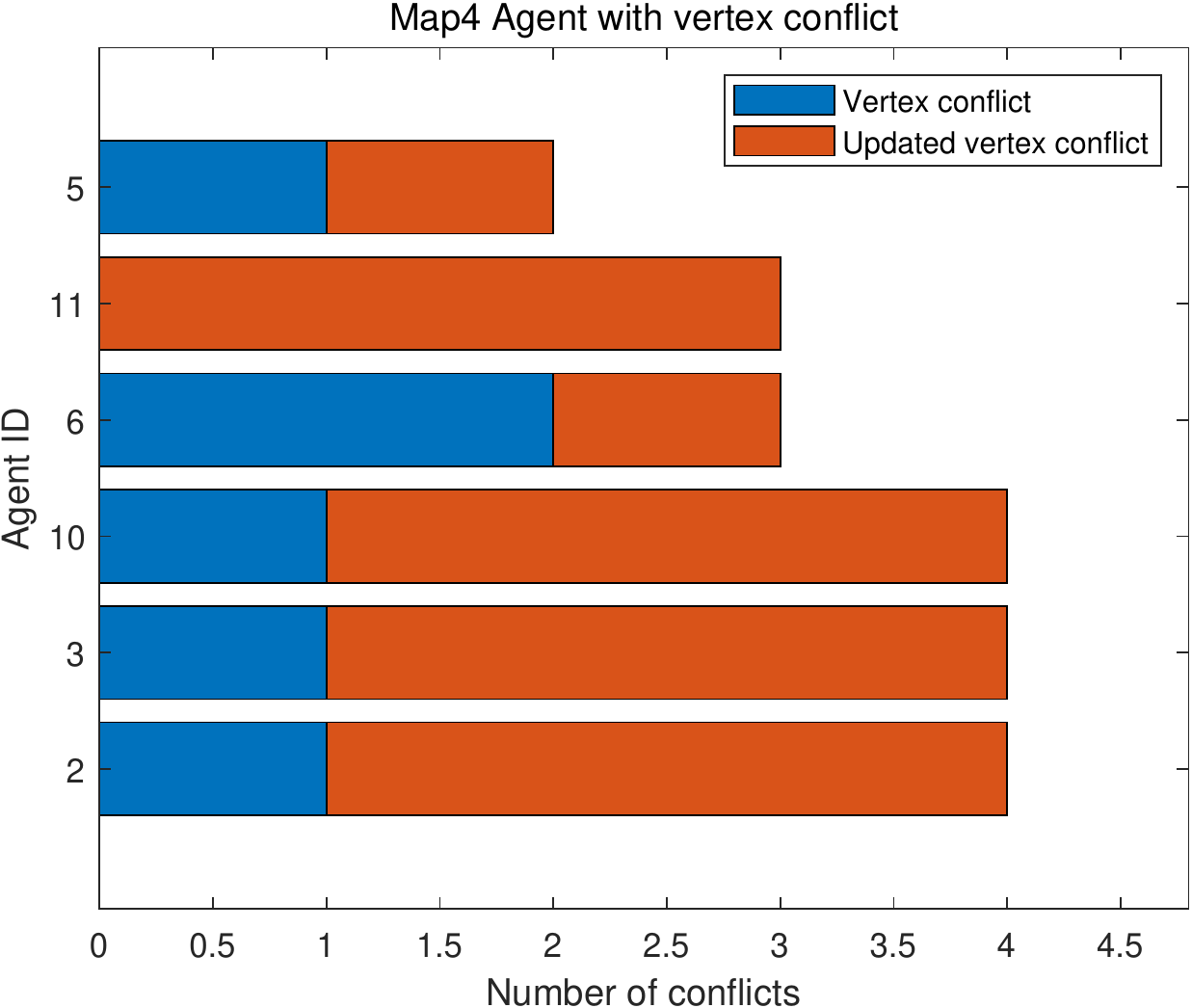}}
            \hfil 
        \caption{Statistics of the conflicts made by corresponding agents on map 4.} \label{fig:TroubleAgentsMap4}
\end{figure*}

In this section, we demonstrate the planned path for each map in Fig. \ref{fig:PathSolution}. As we can see, our algorithm successfully finds out the optimal paths considering the priority of the machines, i.e., the path with the lowest cost considering the main criterion, for all the tested maps.  The algorithm commands the machines to drive directly to the goal, find a bypass, or just wait for others first to pass through.

In Tab. \ref{tab:algorithm duration}, we give the computational time to find out the optimal solution. We divide the searching time into initial search and the following update process. In the first phase, the computational time is no more than 0.1 seconds on our tested maps. In case that the MAPF task is easy, i.e., the counteraction and potential conflicts among the agents are rare, the update process can also be done very fast. As we can see, the total duration to get the optimal solution is within 0.2 seconds for the scenario of agents on map 1. However, in other cases, such as the MAPF tasks on map 2 and map 4, although the period to offer the initial path proposal has no significant difference, the total duration is quite different. Concretely, the tasks on map 2 and map 4 need about 6.8 and 10.4 seconds to be solved. For such tasks, we can of course find the solution inside the BIM system before the machines execute their order saved in the schedule file. In the ideal case, the computational time for finding the solution is not the critical thing. However, in a real application, it is normal that the participants do not act on time when something urgent happens; thus, the ability to replan the path quickly is particularly important rather than let all the machines wait in place. As shown in Fig. \ref{fig:TroubleAgentsMap2} and Fig. \ref{fig:TroubleAgentsMap4}, the update process is a dominant part of the whole searching process. Also, with the data shown in Tab. \ref{tab:algorithm duration}, comparing the duration of the initial search and the following update process, we can conclude that reducing the update process is the main optimization direction to make our algorithm faster.

\begin{table*}[!ht]
	\caption{The average algorithm searching duration (s). ``None'' denote that there is no need to optimize since the response is already quick enough. }
	\centering
	\begin{tabular}{c|ccccccc}
	\hline \hline
	 
     Map Nr.                                          & Map1     & Map2    & Map3    & Map4      & Map5 \\ \hline
      Bidirectional A* algorithm duration & 0.0302   & 0.0844  & 0.0178  &  0.0400   & 0.0322  \\ 
        Update algorithm duration                     & 0.1468   &  6.8192 &  0.1680 & 10.4038   & 1.3689  \\
        Theoretical cost before optimization          & 648.9523 &  1317.0138   & 492.3904    & 435       & 1066  \\
        Total duration by layout optimization         & None     &  0.3058 &  None   & 0.1746    & None\\
        Total duration by agent optimization          & None     &  0.9726 &  None   & 0.1731    & None\\
        Total duration emergence                      & None     & 0.4200  &  None   & 0.1489    & None\\
	\hline \hline
	\end{tabular}
	\label{tab:algorithm duration}
\end{table*}

In this paper, we demonstrate the optimization mechanism on the MAPF task on map 2 and map 4 to avoid wordy; however, we confirm the conclusions we make are also in line with the other maps we tested. The approaches were also tested on some other maps while only the results on map 2 and map 4 are shown in Tab. \ref{tab:algorithm duration}. 
The optimization depends on the stage of the construction site. In the early state before the site is built up in reality, we have more freedom to optimize.  
To accelerate the computation, two methods are proposed for the early stage. The first idea is to modify the unreasonable part of the construction sites. In this fashion, we can improve the throughput of the construction site. Fig. \ref{fig:conflictsPositionMap2} represents the positions where the algorithm commands the machines to pass through but encounters conflicts with other construction machines. As aforementioned, we consider two kinds of conflicts in this study since they are more in line with the construction site, namely edge conflicts and vertex conflicts, respectively.  Fig.  \ref{fig:conflictsPositionMap2} (a) and (c) demonstrate the conflicts found by initial bidirectional search. Since the map is weighted, the best path is usually unique; this is partly proved by the fact that the conflicts number found by initial bidirectional search is a multiple of the time we did the experiments. However, the unique best solution increases the possibility of generating conflicts. Taking the MAPF task on map 2 for example, as shown in Fig. \ref{fig:conflictsPositionMap2}, we can see there are three regions that have more conflicts than others. Concretely,  they are the region including vertex (16, 8) and (17, 8), the region including vertex (2,7), as well as the region including vertex (17,12) and (16,12). Correspondingly, based on the intended movement of the agents around these positions, the algorithm points out that the vertex (16,9), (17,9), (3,8), and (15,12) shall be modified to have similar characteristics as its surrounding. For instance, the road condition of the vertex (16,9) and (17,9) shall be changed into good condition from bad condition since their surroundings have good condition. The computational time is then dramatically reduced to about 0.31 seconds and seems to be the most effective method to reduce the solution finding time. The results are demonstrated in Tab. \ref{tab:algorithm duration}.

The other idea is to remove the most troublemaker in the MAPF task. As we know, the capacity of each construction site has a physical upper limit. No matter how excellent the algorithm is, too many participants will eventually lead to a decline in overall performance. Compared to the first method, which has a potential drawback that there might be some reasons that optimization of the working site is not always feasible, removing a conflicts-causing agent can be used whenever needed. In the case of the MAPF task on map 2, we can remove agent 16 according to Fig. \ref{fig:TroubleAgentsMap2} so that the computational duration reduces to roughly 0.97 seconds. Considering the holistic productivity is only marginally affected since there are still 49 machines that can work well in the working site, and the shorter duration endows the whole system the capability to deal with the emergence on site, this method is recommended if the construction site cannot be modified.

\begin{table*}[!ht]
	\caption{The schedule of MAPF task on map 2. In this task, the algorithm should plan the path for 50 agents.}
	\centering
	\begin{tabular}{c|cccccccccccc}
	\hline \hline
	 
    agent name &  start & 1 & 2 & 3 & 4 & 5 & 6 & 7 & 8 & 9 & 10  \\ \hline
        agent1 & [0, 12] & [1, 12] & [2, 12] & [2, 11] & [2, 10] & [3, 10] & [4, 10] &  &  &  &  &  \\ 
        agent2 & [2, 12] & [2, 11] & [3, 11] & [4, 11] &  &  &  &  &  &  &  &  \\ 
        agent3 & [6, 12] & [7, 12] & [8, 12] & [9, 12] & [10, 12] &  &  &  &  &  &  &  \\ 
        agent4 & [17, 12] & [16, 12] & [16, 12] & [16, 11] & [15, 11] & [15, 10] & [15, 9] &  &  &  &  &  \\ 
        agent5 & [18, 12] & [17, 12] & [17, 12] & [16, 12] & [16, 11] & [15, 11] & [15, 10] &  &  &  &  &  \\ 
        agent6 & [14, 11] & [15, 11] & [16, 11] & [17, 11] & [18, 11] &  &  &  &  &  &  &  \\ 
        agent7 & [0, 10] & [1, 10] & [2, 10] & [2, 9] &  &  &  &  &  &  &  &  \\ 
        agent8 & [8, 10] & [8, 10] & [9, 10] & [9, 9] & [10, 9] &  &  &  &  &  &  &  \\ 
        agent9 & [10, 10] & [9, 10] & [9, 9] & [8, 9] &  &  &  &  &  &  &  &  \\ 
        agent10 & [0, 9] & [1, 9] & [2, 9] & [2, 8] & [2, 7] & [3, 7] &  &  &  &  &  &  \\ 
        agent11 & [2, 9] & [2, 9] & [2, 8] & [2, 7] & [3, 7] & [4, 7] &  &  &  &  &  &  \\ 
        agent12 & [6, 9] & [6, 10] & [7, 10] & [8, 10] &  &  &  &  &  &  &  &  \\ 
        agent13 & [12, 9] & [12, 8] & [12, 7] &  &  &  &  &  &  &  &  &  \\ 
        agent14 & [19, 9] & [19, 8] & [18, 8] & [17, 8] & [16, 8] & [15, 8] & [14, 8] &  &  &  &  &  \\
        agent15 & [5, 8] & [6, 8] & [7, 8] & [7, 7] & [7, 6] & [8, 6] & [9, 6] & [10, 6] & [10, 7] & [10, 8] & [11, 8] &  \\ 
        agent16 & [15, 8] & [16, 8] & [16, 7] & [17, 7] & [17, 8] & [17, 9] &  &  &  &  &  &  \\ 
        agent17 & [19, 8] & [18, 8] & [17, 8] & [16, 8] & [15, 8] & [14, 8] & [13, 8] &  &  &  &  &  \\ 
        agent18 & [0, 7] & [1, 7] & [2, 7] & [2, 6] & [2, 5] &  &  &  &  &  &  &  \\ 
        agent19 & [1, 7] & [2, 7] & [3, 7] & [4, 7] & [4, 6] & [4, 5] & [4, 4] & [3, 4] &  &  &  &  \\ 
        agent20 & [6, 7] & [7, 7] & [7, 6] & [7, 5] &  &  &  &  &  &  &  &  \\ 
        agent21 & [7, 7] & [7, 6] & [8, 6] & [9, 6] & [9, 5] &  &  &  &  &  &  &  \\ 
        agent22 & [13, 7] & [13, 6] & [14, 6] & [14, 5] &  &  &  &  &  &  &  &  \\ 
        agent23 & [15, 7] & [15, 6] & [15, 5] &  &  &  &  &  &  &  &  &  \\ 
        agent24 & [17, 7] & [17, 6] & [17, 5] & [16, 5] &  &  &  &  &  &  &  &  \\ 
        agent25 & [4, 6] & [5, 6] & [6, 6] & [6, 5] &  &  &  &  &  &  &  &  \\ 
        agent26 & [6, 6] & [6, 5] & [6, 4] &  &  &  &  &  &  &  &  &  \\
        agent27 & [1, 0] & [2, 0] & [3, 0] & [4, 0] &  &  &  &  &  &  &  &  \\
        agent28 & [7, 1] & [8, 1] & [8, 0] &  &  &  &  &  &  &  &  &  \\
        agent29 & [13, 0] & [14, 0] & [15, 0] & [16, 0] & [17, 0] & [18, 0] & [19, 0] &  &  &  &  &  \\ 
        agent30 & [17, 6] & [17, 5] & [18, 5] & [19, 5] & [19, 4] &  &  &  &  &  &  &  \\
        agent31 & [19, 6] & [19, 5] & [19, 4] & [19, 3] &  &  &  &  &  &  &  &  \\
        agent32 & [1, 5] & [0, 5] & [0, 4] &  &  &  &  &  &  &  &  &  \\ 
        agent33 & [4, 5] & [4, 4] & [4, 3] &  &  &  &  &  &  &  &  &  \\ 
        agent34 & [7, 5] & [8, 5] & [8, 4] & [8, 3] &  &  &  &  &  &  &  &  \\
        agent35 & [8, 5] & [9, 5] & [10, 5] & [10, 4] &  &  &  &  &  &  &  &  \\
        agent36 & [13, 5] & [12, 5] & [11, 5] & [11, 4] &  &  &  &  &  &  &  &  \\
        agent37 & [17, 5] & [17, 4] & [16, 4] & [16, 3] &  &  &  &  &  &  &  &  \\
        agent38 & [6, 4] & [6, 3] & [6, 2] &  &  &  &  &  &  &  &  &  \\
        agent39 & [12, 4] & [12, 3] & [11, 3] &  &  &  &  &  &  &  &  &  \\
        agent40 & [14, 4] & [14, 5] & [13, 5] & [12, 5] & [12, 4] & [12, 3] &  &  &  &  &  &  \\
        agent41 & [19, 4] & [18, 4] & [17, 4] &  &  &  &  &  &  &  &  &  \\ 
        agent42 & [0, 3] & [0, 2] & [1, 2] & [2, 2] &  &  &  &  &  &  &  &  \\ 
        agent43 & [8, 3] & [8, 2] & [9, 2] & [9, 1] &  &  &  &  &  &  &  &  \\
        agent44 & [15, 3] & [15, 2] & [15, 1] & [14, 1] & [13, 1] &  &  &  &  &  &  &  \\
        agent45 & [18, 3] & [19, 3] & [19, 2] & [19, 1] &  &  &  &  &  &  &  &  \\ 
        agent46 & [0, 2] & [1, 2] & [1, 1] & [2, 1] & [3, 1] & [4, 1] &  &  &  &  &  &  \\
        agent47 & [4, 2] & [4, 1] & [5, 1] & [6, 1] & [6, 0] &  &  &  &  &  &  &  \\
        agent48 & [9, 2] & [10, 2] & [11, 2] & [12, 2] & [13, 2] &  &  &  &  &  &  &  \\
        agent49 & [13, 2] & [14, 2] & [15, 2] & [16, 2] & [17, 2] &  &  &  &  &  &  &  \\
        agent50 & [19, 2] & [18, 2] & [17, 2] & [17, 1] & [16, 1] & [15, 1] &  &  &  &  &  &  \\ 
	\hline \hline
	\end{tabular}
	\label{tab:scheduleMap2}
\end{table*}

In the BIM system, we consciously made sure that the computational duration was within an acceptable period. However, this cannot guarantee all the potential conflicts for these MAPF tasks were removed. For the example of the MAPF task on map 2, in case that agent 16 did not catch up with the planned schedule and had a two-time step delay, the duration for replanning the solution for the whole fleet went to more than 5 seconds, even the construction site was modified in the early stage.  Since we had already optimized the searching process in BIM so that the duration is shorter than 0.5 seconds and the only thing changed here was agent 16, we could quickly draw a conclusion that the expanding computational time was due to agent 16. Afterward, the algorithm checked the path of other agents and found out a new destination for agent 16 to avoid conflicts. Concretely, the new goal for agent 16 shall be the vertex (15,7), and the computational time was then only 0.42 seconds. Notice that agent 16 is not allowed to stop at its original place since it blocks the only way for agent 17 and 14 to reach their goal. In case that machine 16 totally lost its mobility, the algorithm will ask every participant to stop and wait for the human intervention.

In the MAPF task on map 4, we use the same methods to optimize the computation time in order to demonstrate our solution's generalization capability when the terrain is more complex and there are fewer agents. As shown in Fig. \ref{fig:conflictsPositionMap4}, 2 regions have more conflicts than others, i.e., the region including vertex (8,6) and vertex (9,6), and the region including vertex (6,10). Same as the method, namely construction site optimization, used in the previous case, the vertex (5,10) is indicated, which shall be modified from an obstacle to a road with good conditions. In addition, the road condition of the vertex (9,5) and vertex (10,6) shall be changed into good condition from bad condition. After this optimization, the computation time is greatly reduced to roughly 0.17 seconds. For agent optimization, we shall remove the most conspicuous troublemakers in this task, which are agent 10 and agent 11, as shown in Fig. \ref{fig:TroubleAgentsMap4}. Here we removed the agent 10 so that the computation duration reduces to about 0.17 seconds. Although the optimization results of these two methods in the MAPF task on map 4 are almost the same, we recommend layout optimization because only 12 agents were deployed in Map 4. If an agent is removed, the overall productivity will be reduced more significantly compared to the previous scenario on map 2. In case that agent 10 did not run perfectly according to the planned schedule and had a delay of one-time step at the beginning, the computation duration for replanning can also exceed 5 seconds, even if the layout of the construction site was optimized with the first method. Our algorithm gave a new goal for agent 10, which shall be the vertex (7,4), and the computational time was afterward 0.15 seconds. According to the above results, our algorithm is also proven as effective  for the task on map 4.

\section{Advantages of our methods}

In this study, our approach enables many machines to work simultaneously inside of the working site. Firstly, the method helps civil engineers to arrange the construction site before the site is setup. Our approach points out the positions where conflicts occur among the machines and thus indicates the place worthy of being modified. Moreover, our algorithm also helps the engineers to determine the reasonable number of machines in a working site, on the premise of using advanced algorithms. In addition, our algorithm schedules a conflict-free solution for the agents so that the agents can move confidently without hesitation.  Last but not least, since the emergencies are inevitable, we design the system to replan the path solution in a very short period compared to the SOTA MAPF algorithm with only slightly increase the non-optimality of the solution. 

\section{Conclusion}

In this paper, we presented an efficient and effective algorithm to calculate the path of a fleet of machines on a construction site. Considering the complicated terrain of a construction site, we endow our algorithm with the ability to handle the weighted maps. By testing our method on five different and diverse maps, our method successfully found the best path for a fleet including participants with different importance. By solving the MAPF problem for a construction site from both algorithmic and construction layout perspectives, we showed the benefits of our hybrid method, especially in reducing the computational time to handle emergencies. Based on our results, modify the unreasonable part is the most efficient fashion to speed up the searching process. Also, removing the agents which cause the most conflicts is always viable and can dramatically reduce the searching time but slightly reduce the whole productivity.

\subsection{Outlook}
As we mentioned in the literature review, path planning is a fast developing and prosperous research field; thus, we did not fully consider all the improved methods for the initial version of our method. For instance, we did not add in the mega-agents concept, which merges the agents together based on some specific rules to reduce the conflicts and thus speed up the searching process. Since we confirm that our method can be combined with these methods, we expect the searching process to be accelerated further. 

In addition, although Huoshenshan's working site proved the concept, the more machines invested, the faster is the project, which is also consistent with our subjective imagination, a comprehensive study on the quantitative relationship between the number of machines invested and the productivity of the working site is not done. We encourage the experts in civil engineering to propose some challenging scenarios and test our algorithm on them.  

\section{Acknowledgement}

We sincerely thank Dianzhao Li, a PhD student at Technical University of Dresden, for the valuable discussion during his time as a master student at Karlsruhe Institute of Technology  under the supervision from Yusheng Xiang and Prof. Marcus Geimer.

\section{Appendix}

We are happy if you would like to do your research based on this project. However, for commercial use, you should contact us since we have applied for the patents.

\bibliography{IEEE_Literature.bib}{}
\bibliographystyle{IEEEtran}


\newpage
\begin{IEEEbiography}[{\includegraphics[width=1in,height=1.25in,clip,keepaspectratio]{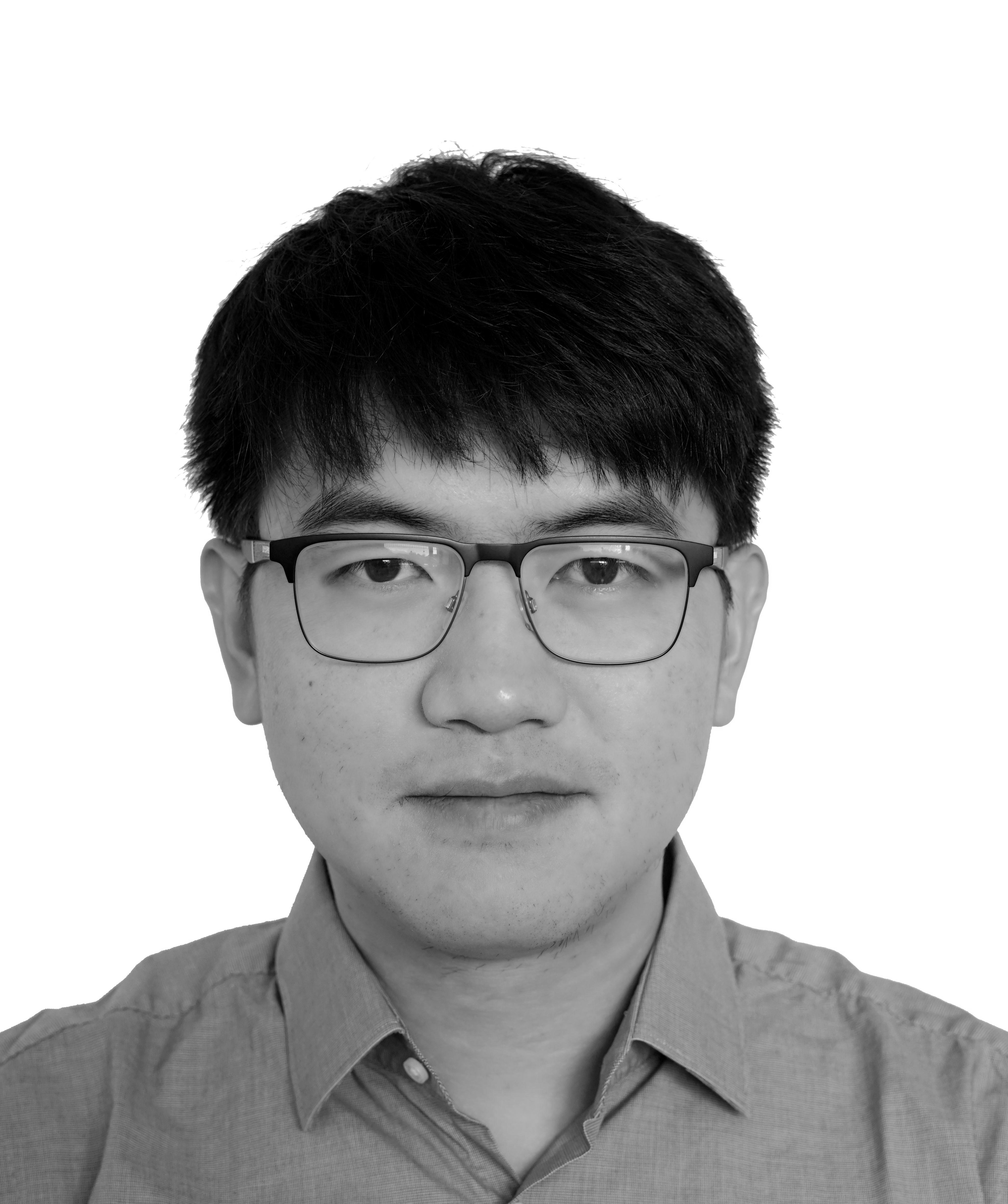}}]{Yusheng Xiang} is pursuing his PhD degree at the Institute of Vehicle System Technology, Karlsruhe Institute of Technology, Karlsruhe, Germany. Also, he was a research scientist at Robert Bosch GmbH, Germany. From Sep. 2020, he is a visiting scholar at the University of California, Berkeley, USA, supervised by Prof Samuel S. Mao. He received M.Sc. degree in Vehicle Engineering with the focus of Mathematical Model Building and Simulation from the Karlsruhe Institute of Technology, Karlsruhe, in 2017. He has authored 9 influential journals and international conference papers, and holds 5 patents. His group deals with the improvement of mobile machines' performance using Artificial Intelligence and the Internet of Things. Currently, he is the CTO of Elephant Tech LLC in Shenzhen, China, which is  a spinoff from Prof. Samuel S. Mao's lab.
\end{IEEEbiography}

\begin{IEEEbiography}[{\includegraphics[width=1in,height=1.25in,clip,keepaspectratio]{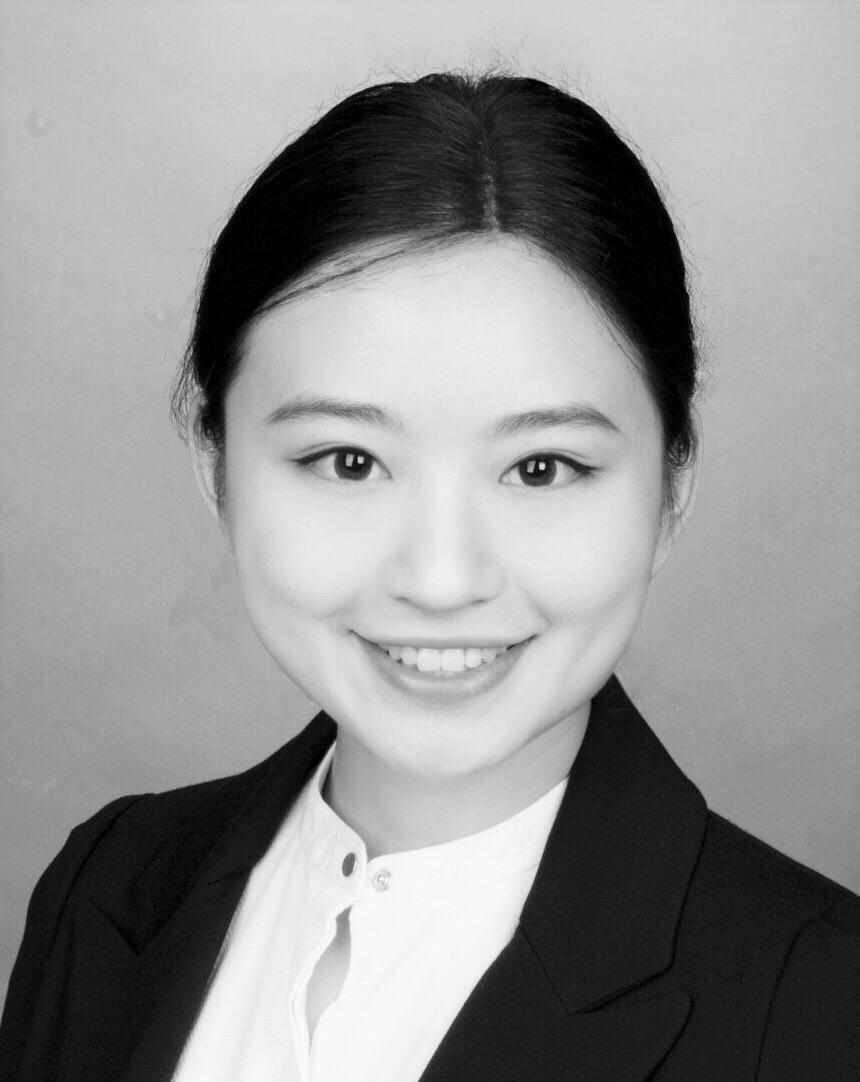}}]{Tianqing Su} is pursuing a full-time Master of Business Administration degree at Guanghua School of Management, Peking University, Beijing, China. She received the B.Sc. degree in telecommunication from the Xidian University, Xi'an, China, in 2013, and the M.Sc. degree in Information Systems Engineering from the Technische Universität Braunschweig, Germany, in 2017.  Before she found the Elephant Tech LLC in Shenzhen and become the first CEO of this hightech startup, she was a Software Engineer in the field of hybrid and electric vehicle at Continental AG, Regensburg, Germany, and Robert Bosch GmbH, Abstatt, Germany. 
\end{IEEEbiography}

\begin{IEEEbiography}[{\includegraphics[width=1in,height=1.25in,clip,keepaspectratio]{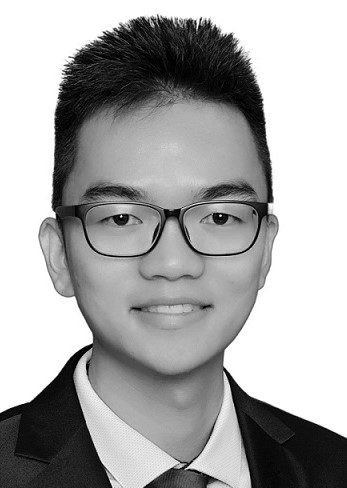}}]{Shirui Ouyang} is pursuing his PhD degree at Institute of Vehicle System Technology, Karlsruhe Institute of Technology, Karlsruhe, Germany. He received M.Sc. degree in Vehicle Engineering with the focus of General Vehicular Technology and Mobile Machines from the Karlsruhe Institute of Technology, Karlsruhe, in 2017, and B.Sc. degree in Vehicle Engineering  from  Beijing Institute of Technology, Beijing,  China. His research focuses on improving the efficiency of mobile working machines by implementation of the hydraulic energy recovery systems and the optimization its control strategy.
\end{IEEEbiography}

\begin{IEEEbiography}[{\includegraphics[width=1in,height=1.25in,clip,keepaspectratio]{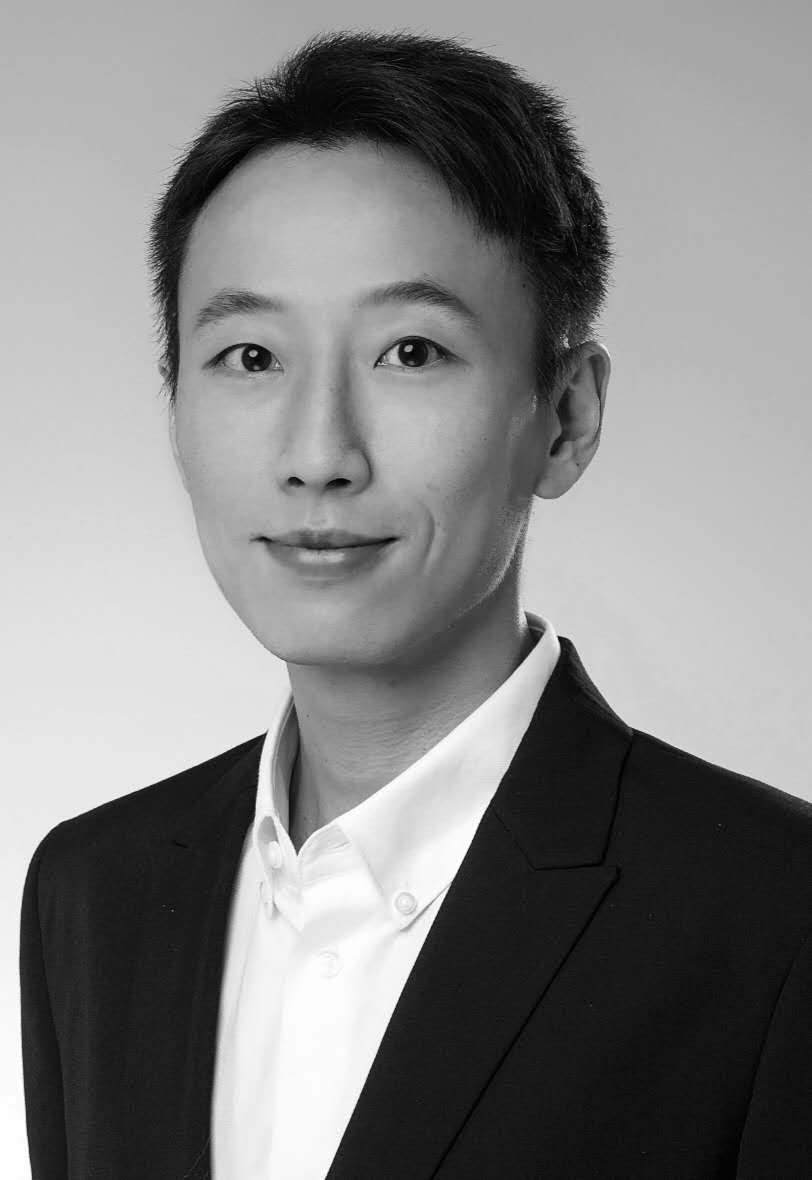}}]{Kailun Liu} received M.Sc. degree in Mechanical Engineering with the focus of Vehicle Technology from the Karlsruhe Institute of Technology, Karlsruhe, Germany, in 2021. He has also received B.Sc. degree in Mechanical Engineering with the focus of Process Equipment and Control Engineering from the East China University of Science and Technology, Shanghai, in 2015. Before he has joined Prof. Geimer's lab to do his master thesis under the supervision from Yusheng Xiang, he did an internship on electric vehicles at Bosch China Central Research Institute in 2019.
\end{IEEEbiography}

\begin{IEEEbiography}[{\includegraphics[width=1in,height=1.25in,clip,keepaspectratio]{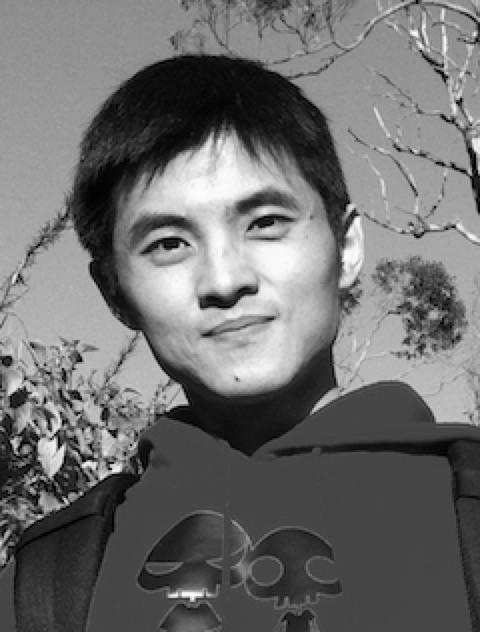}}]{Jun Li}  is a senior lecturer with Artificial Intelligence Institute and School of Computer Science, in the Faculty of Engineering and Information Technology, University of Technology Sydney, Australia. He received his B.S. degree in computer science and technology from Shandong University, China, in 2003, M.Sc. degree in information and signal processing from Peking University, Beijing, China, in 2006, and PhD in computer science from the Queen Mary University of London, U.K., in 2009. His research interest is mostly on probabilistic data models and image and video analysis using neural networks. Currently, he leads the AI department of the Elephant Tech LLC. 
\end{IEEEbiography}

\begin{IEEEbiography}[{\includegraphics[width=1in,height=1.25in,clip,keepaspectratio]{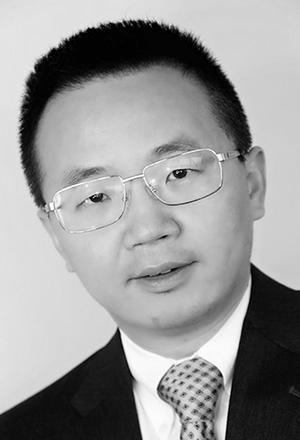}}]{Samuel Mao} received his Ph.D. degree from the University of California at Berkeley in 2000. After that, Prof. Mao started his career at Lawrence Berkeley National Laboratory, where he was a career staff scientist until 2013. He returned to U.C. Berkeley campus as an adjunct professor in 2004, when he also established the Clean Energy Engineering Laboratory that has spun off an international technology development and commercialization institution, the Institute of New Energy, launched in 2013. Having published 180 refereed research articles that have received more than 46,000 citations, Prof. Mao is also an inventor of 80 patents in the U.S. and abroad. He delivered nearly 100 invited, keynote or plenary speeches at international conferences, and previously served as a technical committee member, program review panelist, grant proposal evaluator, and national laboratory observer for the U.S. Department of Energy. In addition to co-founding three international materials and energy technology conferences, he co-chaired Materials Research Society (MRS) annual meeting in the spring of 2011, and the International Conference on Clean Energy in 2012. Prof. Mao received an “R\&D 100” Technology Award (2011) for his technological innovation, and a Berkeley MEGSCO Faculty Teaching Award (2008) for his dedication to higher education.
\end{IEEEbiography}

\begin{IEEEbiography}[{\includegraphics[width=1in,height=1.25in,clip,keepaspectratio]{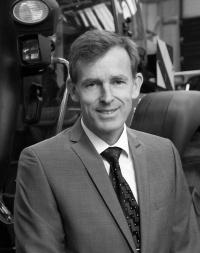}}]{MARCUS GEIMER} received his Diploma degree in mechanical engineering from the RWTH Aachen University, Aachen, Germany in 1990. In 1995, he received his PhD from the Institute of Hydraulics and Pneumatics, today named Institute for Fluid Power Drives and Systems, RWTH Aachen University. He started his industrial career 1995 in the field of construction, where he was the leader of the research group for hydraulic breakers. In 2000, he changed to the hydraulic industry, where he leads the construction and customer development for mobile hydraulics. 
Since 2005, he is a full professor and director at the Institute of Mobile Machines (Mobima), at the Karlsruhe Institute of Technology KIT, Germany. His research activities focus on drives and controls for mobile working machines, like agriculture, construction and municipal vehicles. Research projects on hydrostatic, electric and hybrid drives, as well on traction as on function drives, have been successfully completed. Modern control strategies, like machine learning methods, neural networks or predictive control, are under research. 
\end{IEEEbiography}

\EOD

\end{document}

